\documentclass[10pt,journal,compsoc]{IEEEtran}
\ifCLASSOPTIONcompsoc
  \usepackage[nocompress]{cite}
\else
  \usepackage{cite}
\fi

\ifCLASSINFOpdf
\else
\fi
\usepackage{amsmath,amsfonts,amssymb,amsthm}
\usepackage{algorithm,algorithmic}

  \usepackage{paralist}
  \usepackage{graphics} %% add this and next lines if pictures should be in esp format
  \usepackage{epsfig} %For pictures: screened artwork should be set up with an 85 or 100 line screen
\usepackage{graphicx}
\graphicspath{ {images/} }
\usepackage{caption}
\usepackage{subcaption}
\usepackage{float}
 \usepackage{epstopdf}%This is to transfer .eps figure to .pdf figure; please compile your paper using PDFLaTex or PDFTeXify.
 \usepackage[colorlinks=true,bookmarks=false]{hyperref}
 \usepackage{arydshln}
 \usepackage{multirow}
 \hypersetup{urlcolor=blue, citecolor=red}
\usepackage{hyperref}

\theoremstyle{definition}

\begin{document}
\title{Image retargeting via Beltrami representation}

\author{Chun Pong Lau, %,~\IEEEmembership{Member,~IEEE,}
        Chun Pang Yung and
        Lok Ming Lui}

\markboth{Preprint}
%\markboth{Submitted to Transactions on Image Processing}%
{Shell \MakeLowercase{\textit{et al.}}: Bare Demo of IEEEtran.cls for Computer Society Journals}
\IEEEtitleabstractindextext{%
\begin{abstract}
Image retargeting aims to resize an image to one with a prescribed aspect ratio. Simple scaling inevitably introduces unnatural geometric distortions on the important content of the image. In this paper, we propose a simple and yet effective method to resize an image, which preserves the geometry of the important content, using the Beltrami representation. Our algorithm allows users to interactively label content regions as well as line structures.  Image resizing can then be achieved by warping the image by an orientation-preserving bijective warping map with controlled distortion. The warping map is represented by its Beltrami representation, which captures the local geometric distortion of the map. By carefully prescribing the values of the Beltrami representation, images with different complexity can be effectively resized. Our method does not require solving any optimization problems and tuning parameters throughout the process. This results in a simple and efficient algorithm to solve the image retargeting problem. Extensive experiments have been carried out, which demonstrate the efficacy of our proposed method.
\end{abstract}

% Note that keywords are not normally used for peer review papers.
}

\maketitle

\IEEEdisplaynontitleabstractindextext

\IEEEpeerreviewmaketitle

\IEEEraisesectionheading{\section{Introduction}\label{introduction}}
\IEEEPARstart{I}mage retargeting, which aims to resize an image to one with a prescribed aspect ratio and size, has drawn much attention in recent years. The rapid development of mobile phones and other mobile devices requires images to be displayed with different aspect ratios and sizes, while preserving the structures and details of the important content. It calls for an effective algorithm for image retargeting. 

Motivated by the compelling applications to the problem, lots of studies about image retargeting have been carried out and numerous algorithms have been proposed. Simple scaling is perhaps one of the easiest solutions to the image retargeting problem. However, it often causes unnatural distortions of the important objects in the image, since the method is oblivious to the image content.  Cropping is another simple alternative. Again, as the geometric structures of the salient content are not considered, the method usually causes information loss and cannot produce satisfactory results. To overcome these issues, several techniques have been developed to resize an image in a content-aware fashion. These algorithms take the image content into consideration to preserve important regions and minimize geometric distortions.

Existing approaches for content-aware image retargeting can mainly be divided into two categories, namely, 1. the cropping-based approach and 2. the mesh-based warping approach. Cropping-based methods remove unimportant pixels and shift the remaining pixels in the image to resize the image. For instance, the popular seam carving algorithm proposed by Avidan and Shamir \cite{Avidan2007} searches for seams of minimal importance and remove them to resize the image. Though less important features are removed from the image, artifacts can sometimes be observed and important objects may be broken when seams cut through important regions. Besides, the mesh-based warping approach is another commonly used technique, which is closely related to our work. This approach aims to partition the image by a triangular or quad mesh and resize the image by adjusting the vertex coordinates through a warping map. The geometric distortion of the important region is minimized, while the unimportant region is allowed to have larger distortion. Several warping-based algorithms have been proposed recently. For example, Wolf \cite{Wolf2007} proposed a non-homogeneous content-driven video retargeting algorithm to resize a video. A warping map is obtained by solving a linear system to transform the image that shrinks less important region more than the important one. The important content is automatically detected based on the local saliency, motion detection, and object detectors. Later, Guo \cite{Guo2009} proposed to construct a mesh image representation that is consistent with the underlying image structures, which is then transformed to the target image with a prescribed aspect ratio and size via a stretch minimizing parameterization. Zhang et al. \cite{Zhang2009}  proposed to estimate a warping map by minimizing a quadratic distortion energy that preserves the important regions and image edges. Jin \cite{NHO2010} presented an image resizing algorithm by warping a triangular mesh over the image, which captures the image saliency information as well as the underlying image features. The method can preserve the shapes of curved features in the resized image. Chen \cite{Chen2010} proposed to resize the image by optimizing an energy functional using convex programming on a quadrangular mesh. Panozzo \cite{Panozzo2012} developed a method to warp the image through minimizing an energy functional over the space of axis-aligned deformations subject to different types of constraints, such as the as-similar-as-possible (ASAP) or the as-rigid-as-possible (ARAP) constraints. More recently, Xu et. al\cite{Xu2017} proposed to warp the image using a quasiconformal map, which is obtained by solving an optimization problem.

\begin{figure*}[t]
\centering
\begin{subfigure}[t]{0.28\textwidth}

\includegraphics[width=\textwidth]{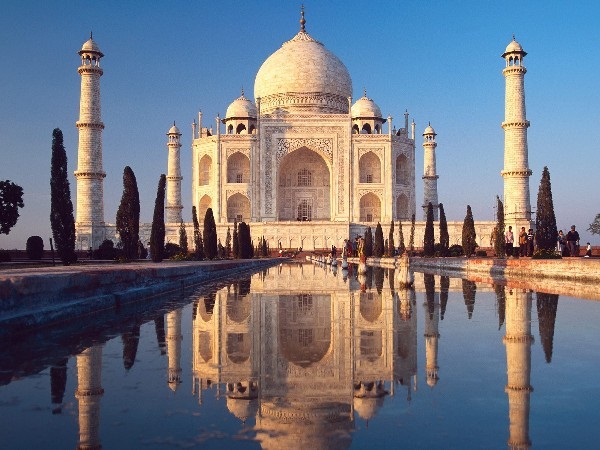}
\caption{}
\end{subfigure}
\begin{subfigure}[t]{0.28\textwidth}
\includegraphics[width=\textwidth]{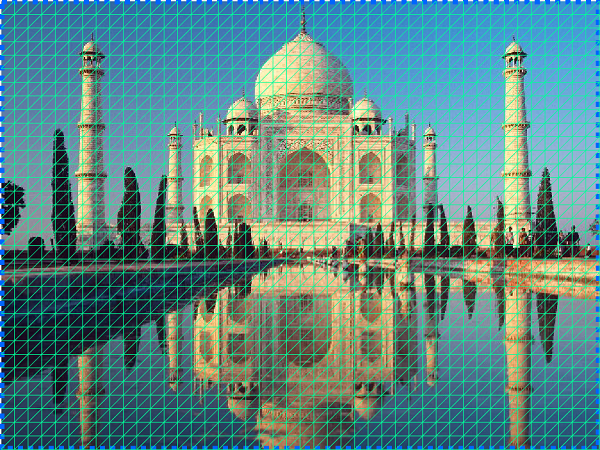}
\caption{}
\end{subfigure} 
\begin{subfigure}[t]{0.28\textwidth}
\includegraphics[width=\textwidth]{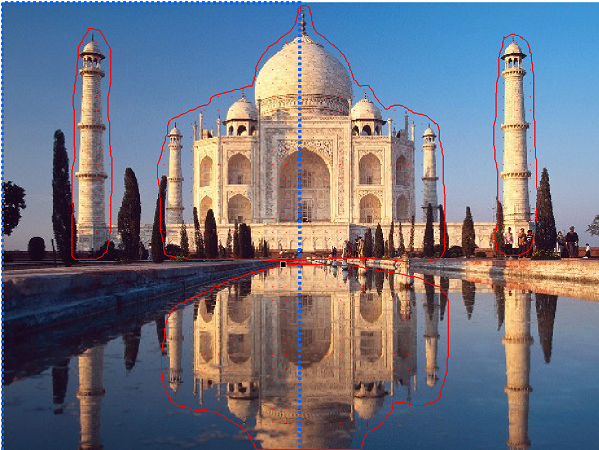}
\caption{}
\end{subfigure}
\begin{subfigure}[t]{0.14\textwidth}
\includegraphics[width=\textwidth]{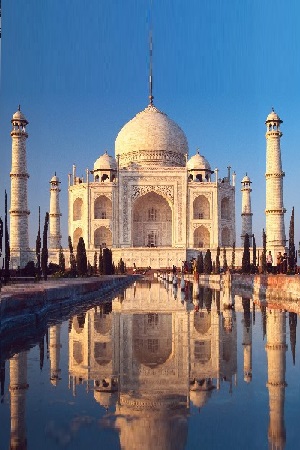}
\caption{}
\end{subfigure}
\caption{Algorithm overview. The image in (a) is firstly overlaid with a regular triangular mesh, as shown in (b). In (c), the important regions are labeled  whose geometry are to be preserved. The aspect ratio of the target image is also set. Finally, the image is resized by adjusting the vertex positions through a warping map, which is shown in (d).}
\label{fig:overview}
\end{figure*}

In this paper, we propose a content-aware image retargeting method that falls into the category of mesh-based warping approaches. We formulate the problem of image retargeting as finding an orientation preserving homeomorphism with controlled distortion to warp an image to a target image of a prescribed aspect ratio, which is content-aware. The warping map is represented by its {\it Beltrami representation}, which is a complex-valued function defined on the image domain. The Beltrami representation captures the local geometric distortion of the warping map. Every warping map is associated with a unique Beltrami representation. By carefully prescribing the values of the Beltrami representation, the associated map with designed distortions at different regions can be reconstructed to warp and resize the image. A resized image that preserves the structures and details of important objects in the image, while minimizing the distortion at the unimportant region, can then be obtained. The use of Beltrami representation is advantageous as it shares the following nice properties to make it fit for the image retargeting problem.

%An effective image targeting method should keep the structures and details of important objects while preserving the homogeneous region with minimal distortion. According to \cite{survey2010}, the image retargeting problem can be formulated as follows. Given an image $I$ of size $m \times n$ and the prescribed size of the target image $m' \times n'$, our goal is to obtain a new image $I'$ of dimension $m' \times n'$ such that the important content and structure are preserved without visual artifacts. In this paper,  we propose a geometric structure-aware method for image retargeting using quasi-conformal maps of triangular meshes. A quasi-conformal map is a generalization of conformal mapping, which is an orientation-preserving bijective mapping. This mapping has a bounded conformality distortion, which intuitively deforms an infinitesimal circle to an infinitesimal ellipse of bounded eccentricity. Every quasi-conformal map is associated with a Beltrami representation, which is a complex-valued Lebesgue measurable function $\mu$ defined on a triangular mesh satisfying $\Vert \mu \Vert_{\infty}<1$. The Beltrami representation measures the local geometric distortion of a deformation. It shares some nice properties to make it useful for the image retargeting problem.

\begin{enumerate}
\item The Beltrami representation effectively measures the local geometric distortion under the warping map. In particular, the local geometry is preserved under the warping map with a zero Beltrami representation. In other words, the geometric distortion under the warping map can be adjusted by manipulating the Beltrami representation.
\item The bijectivity of a map is related to the norm of the Beltrami representation. More specifically, when the supreme norm of the Beltrami representation is strictly less than one, the warping map is guaranteed to be bijective and hence it is foldover-free. A foldover-free warping map is essential for the image retargeting problem so that the warped image has no undesired artifacts.
\item The Beltrami representations of different types of maps can be explicitly formulated. Hence, the Beltrami representation can be easily prescribed without solving any optimization problems to resize an image for different situations. As a result, the computation is fast.
\end{enumerate}

Our proposed image retargeting algorithm using the Beltrami representation is effective for resizing images with different complexity. The method is also simple. It does not require solving any optimization problems and tuning for parameters throughout the process. As a result, the algorithm is efficient. Extensive experiments have been carried out, which demonstrate the efficacy of our proposed model.

\section{Contributions}
The main contributions of this paper are summarized as follows:
\begin{enumerate}
\item An interactive algorithm for image retargeting is proposed, which allows users to label the important regions. The image is deformed by a warping map, whose local geometric distortion can be effectively controlled by the use of the Beltrami representation. Shapes and details of important objects as well as line structures in the image can be preserved by enforcing suitable deformation type constraints.

\item The algorithm does not require solving any optimization problems throughout the process. The computation of the algorithm is efficient.

\item The proposed model is parameter-free so that users do not need to tune for the optimal parameter.

\item The bijectivity of the warping map can be effectively achieved by controlling the norm of the Beltrami representation. This avoids undesired artifacts in the warped image.

\item Three methods to prescribe the values of the Beltrami representation at different regions are presented to resize images with different complexity.
\end{enumerate}

The rest of this paper is organized as follows. In Section \ref{Main}, our proposed content-aware image retargeting method using Beltrami representation is described in details. The numerical implementation of the proposed algorithm is discussed in Section \ref{Numerical implementation}. In Section \ref{experiment}, some experimental results and comparisons with other methods are demonstrated. The paper is concluded in Section \ref{conclusion}.

\begin{figure*}[t]
\centering
\begin{subfigure}[t]{0.24\textwidth}

\includegraphics[width=\textwidth]{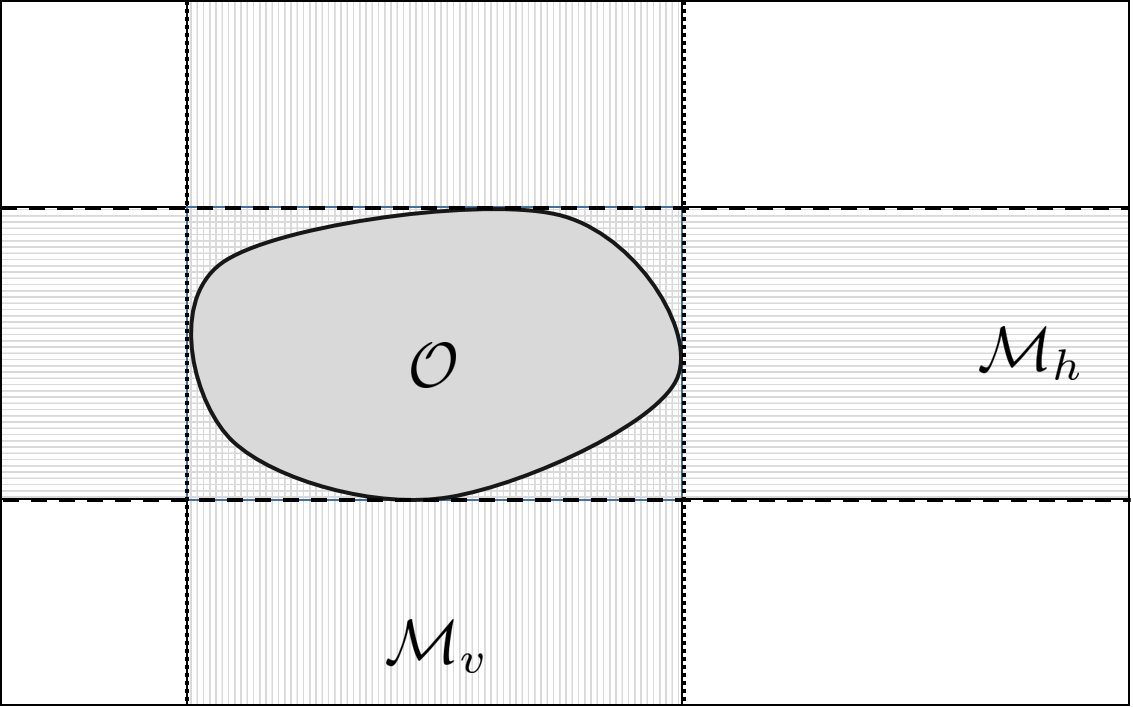}
\caption{}
\end{subfigure}
\begin{subfigure}[t]{0.24\textwidth}
\includegraphics[width=\textwidth]{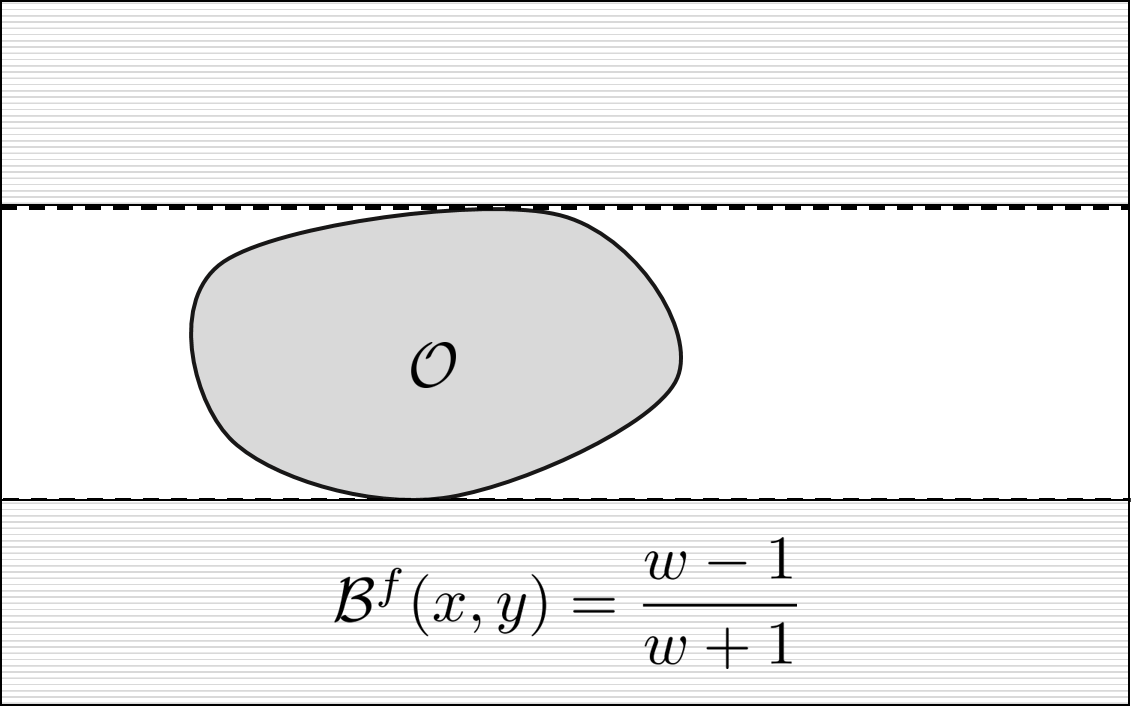}
\caption{}
\end{subfigure} 
\begin{subfigure}[t]{0.24\textwidth}
\includegraphics[width=\textwidth]{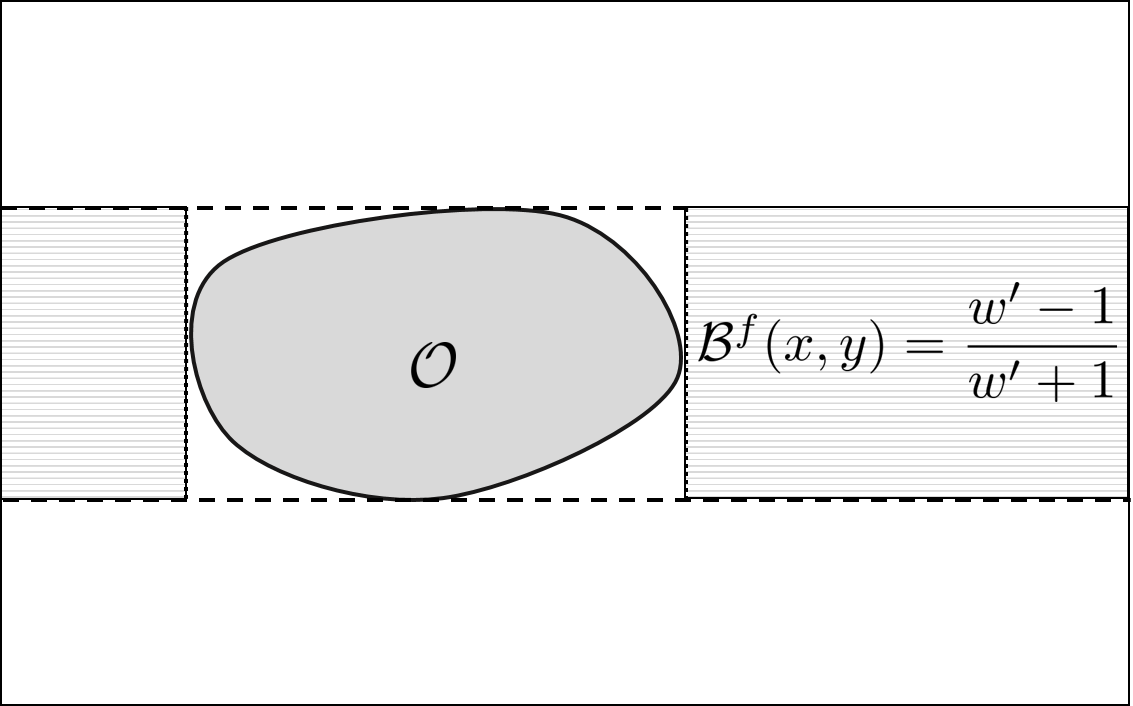}
\caption{}
\end{subfigure}
\begin{subfigure}[t]{0.24\textwidth}
\includegraphics[width=\textwidth]{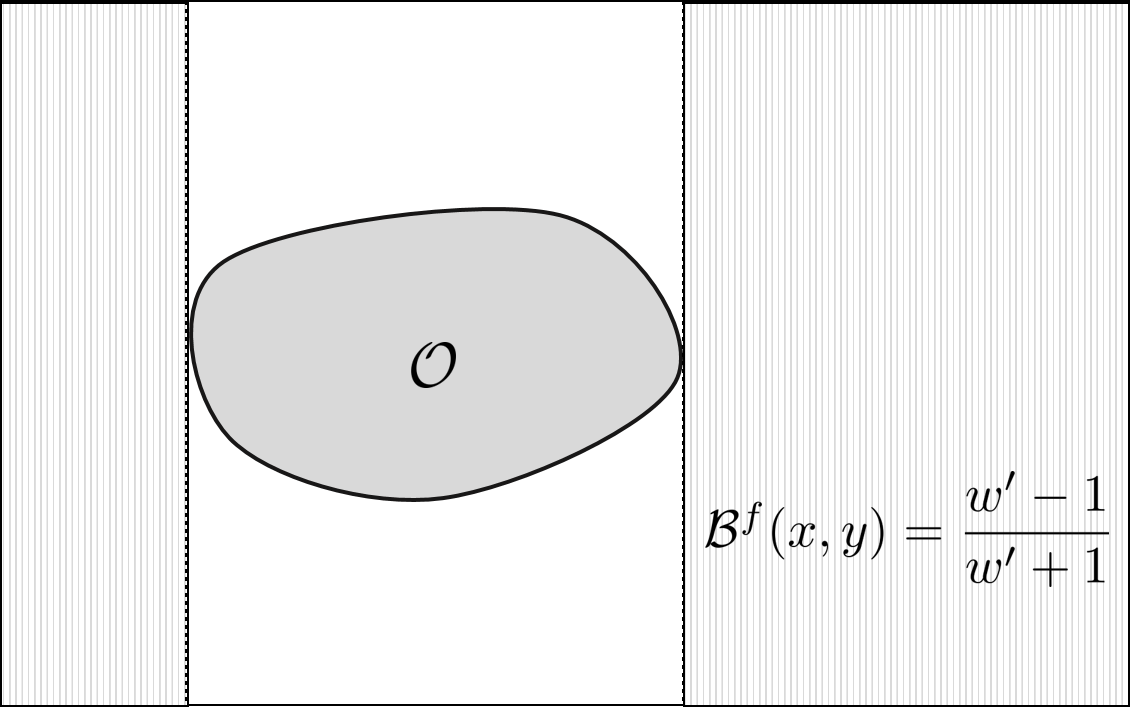}
\caption{}
\end{subfigure}
\caption{Choices of BR. (a) The horizontal and vertical stripes $\mathcal{M}_h$ and $\mathcal{M}_v$. (b) $\&$ (c) Illustration of BR in Choice 2. (d) Illustration of BR in Choice 3. }
\label{fig:BR}
\end{figure*}

\section{Proposed method}\label{Main}
In this section, we describe our proposed image retargeting method using Beltrami representation in details. The main idea is to prescribe suitable values of the Beltrami representation at different regions of the image. An associated warping map can be constructed, which is then applied to warp and resize the image. 

Our problem of content-aware image retargeting can be formally formulated as follows. Suppose an image $I:D\to \mathbb{R}$ is defined on a rectangular domain $D=[0,m]\times [0,n]\subset \mathbb{R}^2$. The target image $I':D'\to \mathbb{R}$ is defined on another rectangular domain $D'=[0,m']\times [0,n']\subset \mathbb{R}^2$ of a different aspect ratio. Denote the regions of important objects and line structures by $\{\mathit{O}_i\}_{i=1}^K$ and $\{\mathit{l}_j\}_{j=1}^L$ respectively. We denote $\mathcal{O}:= \bigcup_{i=1}^K \mathit{O}_i$ and $\mathcal{L}:= \bigcup_{j=1}^L \mathit{l}_j$. Our goal is to find a suitable warping map $f:D\to D'$ such that
\begin{enumerate}
\item $f$ is bijective;
\item $f|_\mathcal{O}$ does not induce geometric distortion in $\mathcal{O}$;
\item The local geometric distortion in $\Gamma:=D\setminus \mathcal{O}$ under $f|_\Gamma$ is small and within an allowable toleration.
\item The line structures of $l_i$ is preserved under $f$.
\end{enumerate}

It is worth noting that the local geometric distortion in $\Gamma$ under $f|_\Gamma$ is inevitable. Condition (2) requires that the geometry is preserved in $\mathcal{O}$ under $f|_\mathcal{O}$. To resize an image to another with a different aspect ratio, all the required geometric distortion must be absorbed in $\Gamma$. Our task is to distribute the geometric distortion nicely over the image such that the warped image looks natural. Our strategy is to use the Beltrami representation to control the distortion under $f$ so that the desirable warping map can be obtained. An overview of our proposed method is summarized in Figure \ref{fig:overview}.

\subsection{Beltrami representation (BR)}
Every warping map $f:D\to D'$ is associated with a unique complex-valued function $\mathcal{B}^f:D\to \mathbb{C}$, called the {\it Beltrami representation (BR)}. Write $f(x,y) = u(x,y) + i v(x,y)$, where $i=\sqrt{-1}$. The BR is defined as follows.
\begin{equation}
\mathcal{B}^f(x,y) = \frac{(\frac{\partial u}{\partial x} - \frac{\partial v}{\partial y})+ i(\frac{\partial v}{\partial x} + \frac{\partial u}{\partial y})}{(\frac{\partial u}{\partial x} + \frac{\partial v}{\partial y})+ i(\frac{\partial v}{\partial x} - \frac{\partial u}{\partial y})} (x,y) \in \mathbb{C} 
\end{equation}
\noindent for all $(x,y) \in D$. Note that BR is defined by the first partial derivatives of $f$, which can be easily discretized on the image domain using the finite difference scheme (see Section \ref{Numerical implementation}). $\mathcal{B}^f (x,y)$ measures the local geometric distortion under $f$ at each point $(x,y) \in D$. In general, a bijective warping map $f$ deforms an infinitesimal circle to an infinitesimal ellipse. The distortion can be measured by $\mathcal{B}^f (x,y)$. From $\mathcal{B}^f (x,y)$, we can determine the angles of the directions of maximal magnification and shrink and the amount of them as well. Specifically, the angle of maximal magnification is $\arg(\mathcal{B}^f (x,y))/2$ with magnifying factor $1 + |\mathcal{B}^f (x,y)|$. The angle of maximal shrinking is the orthogonal angle $(\arg(\mathcal{B}^f (x,y)) − \pi)/2$ with shrinking factor $1 - |\mathcal{B}^f (x,y)|$. Thus, $\mathcal{B}^f$ gives us information about the local geometric distortion under $f$. In particular, the warping map is distortion-free if $|\mathcal{B}^f(x,y)| = 0$ everywhere. The preservation of local geometry under the warping map in the saliency region can be easily achieved by setting the BR to be zero inside the region.

On the other hand, the bijectivity of $f$ can be determined by $\mathcal{B}^f$. It is easy to check that the Jacobian $J(f)$ of $f$ is given by
\begin{equation}
J(f) = [(\frac{\partial u}{\partial x} + \frac{\partial v}{\partial y})^2+ (\frac{\partial v}{\partial x} - \frac{\partial u}{\partial y})^2](1-|\mathcal{B}^f(x,y)|^2).
\end{equation}
As an orientation preserving bijective warping map must have a positive Jacobian everywhere, we conclude that $|\mathcal{B}^f(x,y)|<1$ for every $(x,y)\in D$. Conversely, the associated warping map $f$ is bijective if the magnitude of its BR is everywhere less than 1. In other words, by prescribing a BR whose magnitude is everywhere less than 1, we can obtain a foldover-free warping map. This avoids undesired artifacts in the warped image.

Another useful fact is that the BRs of different types of maps can be explicitly formulated. More specifically, if $f$ is an affine map in a region $\Omega$, then the Beltrami representation of $f$ in that region must be a constant. In other words, $\mathcal{B}^f(z) = k\in \mathbb{C}$ in $\Omega$. Also, if $f$ is a simple scaling map in a region $\Omega$ (scaling the horizontal axis and vertical axis by two different scalars), then the  Beltrami representation of $f$ in that region must be a real number. In particular, if $f(x,y) = (ax, by)$ in $\Omega$, then $\mathcal{B}^f(z) = \frac{a-b}{a+b}\in \mathbb{R}$ in $\Omega$. If $a = b$, the Beltrami representation is equal to 0 in the region. Hence, we can easily design the types of deformations at different regions of the image, in order to obtain a natural retargeted image.

\subsection{Choices of BR}
A warping map is associated to a unique BR that captures the local geometric distortion under the map. To resize an image naturally, we look for a suitable warping map with desirable distortions at different regions. The distortions at different regions can be described effectively by the BR. Our strategy to obtain the suitable warping map is by assigning the values of BR according to the desired types of deformations at different regions. An optimal warping map whose BR is closest to the prescribed one can be constructed. According to various situations, we present three choices of BR to resize images with different complexity.

\subsubsection*{Choice 1: Even distribution of distortions}
Resizing an image can be regarded as stretching or squeezing the image in the horizontal direction. We will discuss the case when the image is squeezed in the horizontal direction. The case when the image is stretched in the horizontal direction can be done similarly by rotating the image by 90 degrees. 
Suppose the width to height ratio of our target image is given by $wm:n$. For simplicity, we assume the height of the target image is always kept as the same as the original image. Then, $w$ is always less than 1. Our goal is to warp the original image to the target image, such that the structures in the important region are preserved under the warping map $f$. This can be achieved by enforcing $f$ to be a uniform scaling map in $\mathcal{O}$. In other words, $f(x,y) = (Ax, Ay)$ for some $A\in \mathbb{R}^+$. The BR of a uniform scaling map is zero. Hence, we set the BR in $\mathcal{O}$ as $\mathcal{B}^f(x,y) = 0$ for $(x,y)\in \mathcal{O}$. To resize the original image to the target image, we can warp the image by simple scaling $\mathcal{S}(x,y) = (wx,y)$. The BR of $\mathcal{S}$ can be explicitly computed, which is given by $\mathcal{B}^{\mathcal{S}}(x,y) = \frac{w-1}{w+1}$. Since the local geometry should be preserved in $\mathcal{O}$, we propose to distribute the distortion in the unimportant region. In other words, we set the BR outside $\mathcal{O}$ as $\mathcal{B}^f(x,y) = \frac{w-1}{w+1}$ for $(x,y)\not\in \mathcal{O}$. Intuitively, we distribute the inevitable distortion evenly over the unimportant region. 

In summary, the BR is prescribed as follows.
\begin{equation}
\mathcal{B}^f(x,y)=
\begin{cases}
0 \ \ \text{ if } (x,y)\in \mathcal{O},\\
\frac{w-1}{w+1} \ \ \text{   if }(x,y)\not\in \mathcal{O}.\\
\end{cases}
\end{equation}

Note that the prescribed BR may not be admissible subject to the constraint of the given aspect ratio. That is, the prescribed BR may not correspond to a warping map that deforms an image to another with the given aspect ratio. In the next subsection, we will introduce an algorithm, called the {\it Constrained Linear Beltrami Solver}, to obtain a warping map whose BR is as close as the prescribed one as possible.

\begin{figure*}[t]
\centering
\begin{subfigure}[t]{0.16\textwidth}
\includegraphics[width=\textwidth]{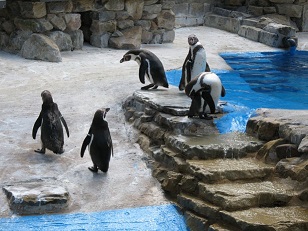}
\caption{}
\end{subfigure}
\begin{subfigure}[t]{0.16\textwidth}
\includegraphics[width=\textwidth]{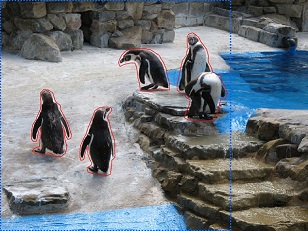}
\caption{}
\end{subfigure} 
\begin{subfigure}[t]{0.16\textwidth}
\includegraphics[width=\textwidth]{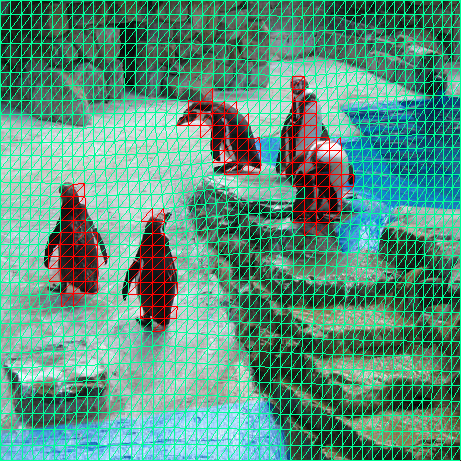}
\caption{}
\end{subfigure}
\begin{subfigure}[t]{0.16\textwidth}
\includegraphics[width=\textwidth]{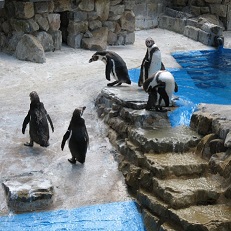}
\caption{}
\end{subfigure}
\begin{subfigure}[t]{0.16\textwidth}
\includegraphics[width=\textwidth]{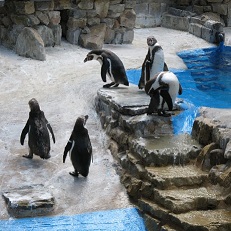}
\caption{}
\end{subfigure}
\begin{subfigure}[t]{0.16\textwidth}
\includegraphics[width=\textwidth]{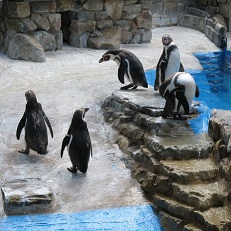}
\caption{}
\end{subfigure}
\caption{Proposed method with Choice 1. (a) Original image. (b) Image with labeled important objects. (c) Resized image to 75\% of the original width using Choice 1. The deformed mesh is also shown. (d) Resized image using Choice 1. (e) Resized image using Choice 2. (f) Resized image using Choice 3.}
\label{fig:mono mu}
\end{figure*}

\subsubsection*{Choice 2: Weakly uneven distribution of distortions}
In Choice 1, the local geometry in $\mathcal{O}$ is preserved under $f$, since the warping map is restricted to be a uniform scaling. However, the sizes of the important objects can still be altered. In order to better keep the sizes of the important objects, we must tolerate more distortion in the unimportant region. 

To describe the assignment of BR in this choice, we denote the horizontal and vertical stripes covering the important objects by $\mathcal{M}_h$ and $\mathcal{M}_v$ respectively. See Figure \ref{fig:BR}(a) for an illustration. Suppose the total width of the important objects $O_i$ is given by $W$. To warp the image so that the sizes of the important objects are kept, the horizontal direction should be scaled more in $\mathcal{M}_h\setminus \mathcal{O}$. In Choice 1, the warping map scales the interval $[0,m]$ to $[0,wm]$. Here, we scale $[0,m]$ to $[0, wm - W]$. Therefore, we prescribe the BR as $\mathcal{B}^f(x,y) = \frac{w'-1}{w'+1}$ for $(x,y)\in \mathcal{M}_h\setminus \mathcal{M}_v$, where $w' = w - \frac{W}{m}$. It is easy to check that $\Big|\frac{w'-1}{w'+1}\Big|> \Big|\frac{w-1}{w+1}\Big|$ for $w<1$. In other words, we are tolerating larger distortion in the unimportant region than that in Choice 1, in order that the sizes of the important objects can be better kept.  In $D\setminus \mathcal{M}_h$, we prescribe the BR as in Choice 1. Under this assignment, the inevitable distortion is unevenly distributed over the unimportant region.

In summary, the BR is prescribed as follows.
\begin{equation}
\mathcal{B}^f(x,y)=
\begin{cases}
0 \ \ \text{ if } (x,y)\in \mathcal{M}_h\cap \mathcal{M}_v,\\
\frac{w'-1}{w'+1} \ \ \text{   if }(x,y)\in \mathcal{M}_h\setminus \mathcal{M}_v,\\
\frac{w-1}{w+1} \ \ \text{   if }(x,y)\in D\setminus \mathcal{M}_h.
\end{cases}
\end{equation}
For an illustration of Choice 2, please refer to Figure \ref{fig:BR}(b) and (c).

\subsubsection*{Choice 3: Strongly uneven distribution of distortions}
In Choice 2, a larger distortion is assigned to the unimportant region covered by the horizontal stripes and line structures. The sizes of the important objects can be less distorted. To further keep the sizes of the important objects, we should assign a larger distortion over a larger area of the unimportant region. In Choice 3, we assign a larger distortion over the unimportant region that is not covered by the vertical stripes and line structures. That is, $D\setminus \mathcal{M}_v$. More specifically, we assign the BR in this area as $\mathcal{B}^f(x,y) = \frac{w'-1}{w'+1}$ for $(x,y)\in D\setminus \mathcal{M}_v$. This area is larger than that of Choice 2. In other words, a stronger distortion is distributed over a larger area of the unimportant region. As a result, the sizes of the important objects can be better kept. On the other hand, the transition of BR values in the vertical stripes from the important objects to the unimportant region causes an unnatural change in distortion. The unnatural transition of local geometric distortions can result in the unnatural deformation of the overall image. To solve this issue, we set the warping map to be a uniform scaling over the vertical stripes. Thus, the BR in $\mathcal{M}_v$ is prescribed as $\mathcal{B}^f(x,y) = 0$ for $(x,y)\in \mathcal{M}_v$. Under this assignment, the inevitable distortion is more unevenly distributed over the unimportant region than the previous choice.

In summary, the BR is prescribed as follows.
\begin{equation}
\mathcal{B}^f(x,y)=
\begin{cases}
0 \ \ \text{ if } (x,y)\in \mathcal{M}_v,\\
\frac{w'-1}{w'+1} \ \ \text{   if }(x,y)\in D\setminus \mathcal{M}_v.\\
\end{cases}
\end{equation}
For an illustration of Choice 3, please refer to Figure \ref{fig:BR}(d).

\subsubsection*{Extremal situation}
Choice 2 and Choice 3 assume the image after resizing can still enclose the important objects of the same sizes. In the extremal situation when the aspect ratio is very small, the image has to be squeezed and cannot enclose all the important objects with the same sizes. In this case, $w'$ has to be carefully set. We assume the total width $W$ of the important objects occupy $\beta$ percents of the width of the target image. $w'$ is then set as $w' = w(1-\beta)/200$. Since the sizes of the important objects are to be resized, the height should also be scaled. The scaling ratio $h$ is given by $h = (n-Hw')/(n-H)$, where $H$ is the total height of the important objects. Therefore, in Choice 2, the BR is prescribed as $\frac{w'-h}{w'+h}$ in $\mathcal{M}_h\setminus \mathcal{M}_v$ and $\frac{w-h}{w+h}$ in $D\setminus \mathcal{M}_h$.

\subsection{Warping map from prescribed BR}
After the BR is prescribed, we proceed to construct an associated warping map. Given the prescribed aspect ratio, the prescribed BR may not correspond to a warping map that deforms an image to another with the given aspect ratio. Our goal is to look for a warping map whose BR is as close as the prescribed one as possible.

If $\mathcal{B}^f=\rho+i \tau$ is the BR of $f = u+iv$, it satisfies the following partial differential equation:
\begin{equation}
 D_1 f = \mathcal{B}^f D_2 f,
\end{equation}
where $ D_1 f = (\frac{\partial u}{\partial x} - \frac{\partial v}{\partial y})+ i(\frac{\partial v}{\partial x} + \frac{\partial u}{\partial y})$ and $ D_2 f= (\frac{\partial u}{\partial x} + \frac{\partial v}{\partial y})+ i(\frac{\partial v}{\partial x} - \frac{\partial u}{\partial y})$. From the above equation, we can check that $f$ also satisfies the following elliptic PDEs: 
\begin{equation}\label{equation:elliptic PDE}
\nabla \cdot  \bigg(A \nabla f\bigg) = 0,
\end{equation}
\noindent where $A=\begin{pmatrix}
\alpha_1 & \alpha_2 \\
\alpha_2 & \alpha_3
\end{pmatrix}$ and
$\begin{cases}
\alpha_1=\dfrac{(\rho-1)^2+\tau^2}{1-\rho^2-\tau^2}; \\
\alpha_2=-\dfrac{2\tau}{1-\rho^2-\tau^2}; \\
\alpha_3=\dfrac{(\rho+1)^2+\tau^2}{1-\rho^2-\tau^2}.
\end{cases}$

To retarget an image naturally, some constraints must be imposed on the warping map. 

\subsubsection*{A. Uniform scaling in the object regions}
Firstly, in order to preserve the geometric structures of the important objects, we require that the warping map to be a uniform scaling map in each object region $O_i$. Mathematically, we impose that:
\begin{equation}
f|_{O_i}(\overrightarrow{x}) = r_o \overrightarrow{x}+\overrightarrow{t_{o_i}}
\end{equation}
\noindent for some scaling parameter $r_o\in \mathbb{R}^+$ and translation vector $\overrightarrow{t_{o_i}}\in \mathbb{R}^2$. Note that the scaling factor $r_o$ is chosen to be the same for every object regions. The translation vector $\overrightarrow{t_{o_i}}$ is to be determined for each $O_i$.

Similarly, we impose that the warping map $f$ on the line structure $l_j$ is of the following form:
\begin{equation}
f|_{l_j}(\overrightarrow{x}) = (r_{l_j}^x x, r_{l_j}^y y) +\overrightarrow{t_{l_j}}
\end{equation}
\noindent where $\overrightarrow{x} = (x,y)$ for some scaling parameter $\overrightarrow{r_{l_j}} = (r_{l_j}^x , r_{l_j}^y) \in \mathbb{R}^+ \times \mathbb{R}^+$ and translation vector $\overrightarrow{t_{l_j}}\in \mathbb{R}^2$, which are to be determined.  Note that the scaling factor $\overrightarrow{r_{l_j}}$ and the translation vector $\overrightarrow{t_{l_j}}$ are determined for each $l_j$.

Since a uniform scaling map is conformal, it is consistent with our choice of Beltrami representation in $\Omega$, which is set to be zero. In summary, our desired quasi-conformal warping map $f$ should satisfy 
\begin{equation} \label{eq:line constraint}
    \begin{cases}
        \nabla \cdot A \nabla (f) =0 \\
        f(\partial D)= \partial D' \\
        f\big|_{O_i} (\overrightarrow{x})= r_o \overrightarrow{x} + \overrightarrow{t_{o_i}} \quad \forall i\\
        f\big|_{\mathit{l}_j} (\overrightarrow{x})= \overrightarrow{r_{l_j}}\cdot \overrightarrow{x} + \overrightarrow{t_{l_j}} \quad \forall j
    \end{cases}
\end{equation}
\noindent for some $r_o \in \mathbb{R}^+$, $\overrightarrow{r_{l_j}} \in \mathbb{R}^+ \times \mathbb{R}^+$ and $\overrightarrow{t_{o_i}}, \overrightarrow{t_{l_j}}\in \mathbb{R}^2$, which are to be determined from the proposed algorithm.

The elliptic PDE subject to the above constraints can be discretized into a sparse linear system. Note that the scaling parameters and translation vectors are not chosen in advance. These parameters will be included as unknown variables in the linear system, which are to be determined. We call this solver for the warping map as the {\it Constrained Linear Beltrami Solver (CLBS)}. The details of the discretization can be found in the next section. 

\bigskip

\noindent {\bf Remark:} {\it Note that the scaling parameters and translation vectors have to be chosen carefully such that the important region will not be expanded or squeezed exceedingly. An extreme squeezing may result in an unnatural resized image, while the important region cannot be expanded outside the boundary of the target domain. As the Beltrami representation $\mathcal{B}^f(x,y)$ satisfies $\Vert \mathcal{B}^f(x,y) \Vert < 1$, it ensures that the scaling parameters and translation vectors can be determined suitably. The scaling and translation of the important region are determined to yield a natural resized image.}

\bigskip

\subsubsection*{B. Preserving line structures in the background}
On the other hand, some images may contain horizontal and vertical line structures in the background near the important objects. Visible distortions of these line structures may result in a visually unnatural resized image. In order to tackle this situation, we propose a special constraint, called the {\it Chessboard constraint}, to better preserve these linear structures. More specifically, we impose the constraint on the type of deformation in $\mathcal{M}_h$ and $\mathcal{M}_v$ as follows.
\begin{equation}
\begin{split}
    f\big|_{\mathcal{M}_h}(\overrightarrow{x})= r \overrightarrow{x} + \overrightarrow{t_i^x};\\
    f\big|_{\mathcal{M}_v}(\overrightarrow{x})= r \overrightarrow{x} + \overrightarrow{t_i^y};\\
\end{split}
\end{equation}

Again, the scaling parameter and translation vectors are to be determined and not prescribed in advance. We impose the same scaling parameter $r$ for both $\mathcal{M}_h$ and $\mathcal{M}_v$. $\overrightarrow{t_i^x}$ is required to be a horizontal translation, that is, $\overrightarrow{t_i^x} = (c_i^x,0)$ for some constant $c_i^x\in \mathbb{R}$. Similarly, $\overrightarrow{t_i^y}$ is required to be a vertical translation, that is, $\overrightarrow{t_i^y} = (0,c_i^y)$ for some constant $c_i^y\in \mathbb{R}$. These constraints ensure the horizontal structures are kept to be horizontal and vertical structures are kept to be vertical. In summary, our desired warping map $f$ should satisfy:
\begin{equation} \label{eq:line constraint discrete}
    \begin{cases}
        \nabla \cdot A \nabla (f) =0 \\
        f(\partial D)= \partial D' \\
        f\big|_{\mathcal{M}_h} (\overrightarrow{x})= r \overrightarrow{x} + \overrightarrow{t_i^x} \quad \forall i \\
        f\big|_{\mathcal{M}_v} (\overrightarrow{x})= r \overrightarrow{x} + \overrightarrow{t_i^y} \quad \forall i 
    \end{cases}
\end{equation}
\noindent for some $r\in \mathbb{R}^+$ and $\overrightarrow{t_i^x}, \overrightarrow{t_i^y}\in \mathbb{R}^2$, which are to be determined from the proposed algorithm. Again, the constrained elliptic PDE can be discretized into a linear system and solved by CLBS.

\begin{figure*}[t]
\centering
\begin{subfigure}[t]{0.38\textwidth}
\includegraphics[width=\textwidth]{penguins}
\caption{}
\end{subfigure}
\begin{subfigure}[t]{0.285\textwidth}
\includegraphics[width=\textwidth]{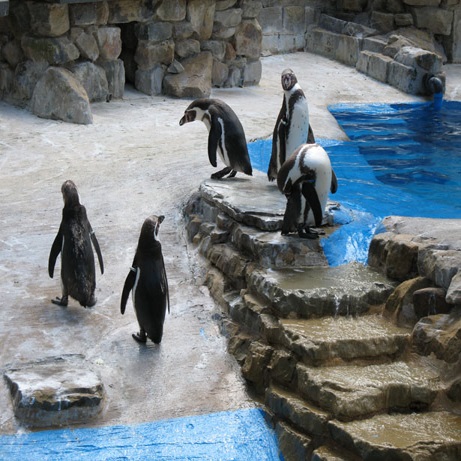}
\caption{}
\end{subfigure} 
\begin{subfigure}[t]{0.285\textwidth}
\includegraphics[width=\textwidth]{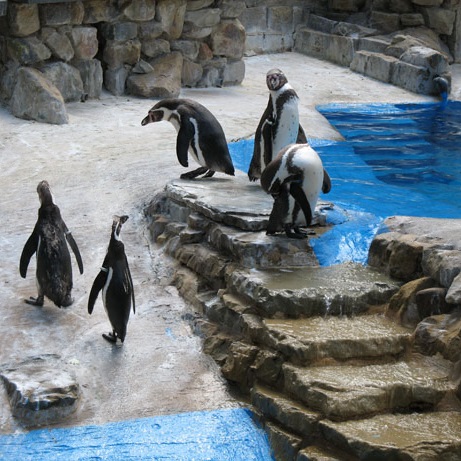}
\caption{}
\end{subfigure} \\
\begin{subfigure}[t]{0.285\textwidth}
\includegraphics[width=\textwidth]{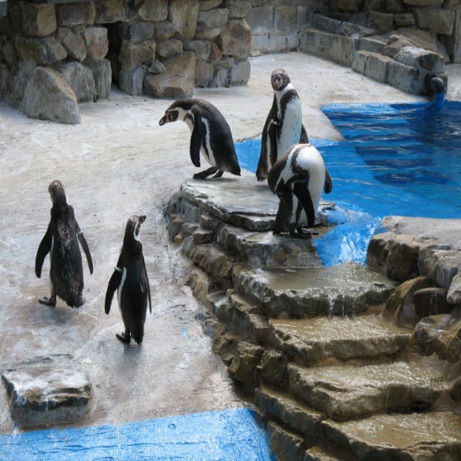}
\caption{}
\end{subfigure}
\begin{subfigure}[t]{0.285\textwidth}
\includegraphics[width=\textwidth]{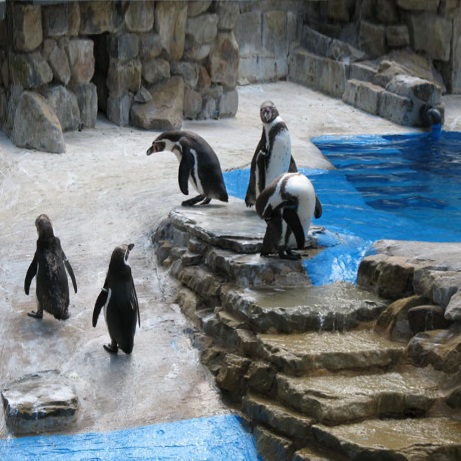}
\caption{}
\end{subfigure}
\begin{subfigure}[t]{0.285\textwidth}
\includegraphics[width=\textwidth]{penguins_mono_line}
\caption{}
\end{subfigure}
\caption{Comparison of different methods to resize a penguin image to 75\% of the original width. (a) Original image. (b) Simple scaling. (c) Seam-Carving (SC). (d) Nonhomogeneous warping (Warp). (e) Scale-and-Stretch (SNS). (f) Proposed method.}
\label{fig:penguin}
\end{figure*}

\section{Implementation} \label{Numerical implementation}
In this section, we describe the discretization and numerical implementation of our proposed model in details.

Suppose a rectangular image $I$ is of size $m \times n$ and the target image $I'$ is of size $m' \times n'$. Let $\mathcal{K}(I)$ be a regular triangular mesh discretizing $I$. Denote the vertices and faces of $\mathcal{K}(I)$ by $\{ V_i =(x_i,y_i) : i=1,\dots,n \}$ and $T_I=[V_i,V_j,V_k]$ respectively. With the warping map, $\mathcal{K}(I)$ is transformed to another mesh $\mathcal{K}(I')$ discretizing the target image $I'$. We denote the vertices and faces of $\mathcal{K}(I')$ by $\{ V'_i : i=1,\dots,n \}$ and $T_I'=[V'_i,V'_j,V'_k]$ respectively. Note that there is a 1-1 correspondence between $V_i$'s and $V'_i$'s.

To represent the line structure $l_j$, the intersected faces of $\mathcal{K}(I)$ with the lines are landmarked. The collection of these faces are denoted by $\mathcal{K}(l_j)$. For the object region, the contour enclosing the region will be delineated. The intersected faces of $\mathcal{K}(I)$ with the contour as well as the interior faces inside the contour are landmarked. The collection of these faces are denoted by $\mathcal{K}(O_i)$.

Given the prescribed Beltrami representation, a discrete quasi-conformal map is computed with suitable constraints to resize the image with the given aspect ratio. In this paper, we use a modified version of the Linear Beltrami Solver (LBS)\cite{Lui2013}\cite{Lam14} with additional constraints. The LBS numerically solves the elliptic PDEs (\ref{equation:elliptic PDE}) by discretizing the generalized Laplacian operator $\nabla \cdot A \nabla$, where $A$ is as defined in equations (\ref{equation:elliptic PDE}). 

In the discrete formulation, a warping map $f$ is a simplicial map, which is piecewise linear on each triangular face $T$. Denote the Beltrami representation by $\mathcal{B}^f=\rho+i \tau$, where both $\rho$ and $\tau$ are complex-valued function defined on each face. Denote the associated discrete warping map by $f = u+ iv$. Then, on each face $T$, $f$ can be written as follows:
\begin{equation}
    f|_T(x,y)=
    \begin{pmatrix}
        u|_T(x,y) \\
        v|_T(x,y)
    \end{pmatrix}=
    \begin{pmatrix}
        a_T x + b_T y + r_T \\
        c_T x + d_T y + s_T
    \end{pmatrix}
\end{equation}
\noindent for some real constants $a_T,b_T,c_T,d_T,r_T$ and $s_T$. Note that the first derivative of $u$ and $v$ on each face $T$ are given by:
\begin{equation}
    u_x|_T=a_T,  u_y|_T=b_T, v_x|_T=c_T,  v_y|_T=d_T
\end{equation}

Suppose faces $T$ and $f(T)$ have vertexes ${[v_{i},v_{j},v_{k}]}$ and ${[w_{i},w_{j},w_{k}]}$ respectively. The edges $\overrightarrow{v_i v_j}=v_j-v_i$ and $\overrightarrow{v_i v_k}=v_k-v_i$, $f$ should then be mapped to $\overrightarrow{w_i w_j}=w_j-w_i$ and $\overrightarrow{w_i w_k}=w_k-w_i$ respectively, that is,

\vspace{-6mm}

\begin{equation}
    \begin{pmatrix}a_{T} & b_{T}\\ c_{T} & d_{T} \end{pmatrix}\begin{pmatrix}g_{j}-g_{i} & g_{k}-g_{i}\\ h_{j}-h_{i} & h_{k}-h_{i} \end{pmatrix} \\ =\begin{pmatrix}s_{j}-s_{i} & s_{k}-s_{i}\\ t_{j}-t_{i} & t_{k}-t_{i} \end{pmatrix}
\end{equation}
\noindent where $v_n=g_n + i h_n$ and $w_n = s_n + i t_n$.

It is easy to check that $det \begin{pmatrix}a_{T} & b_{T}\\ c_{T} & d_{T} \end{pmatrix} $ the signed area of the parallelogram, which is ${2 Area(T)}$. As a result, we have 
\begin{align*} &\quad \begin{pmatrix}a_{T} &  b_{T}\\
 c_{T} & d_{T}\end{pmatrix}  \\
 &=  \frac{1}{2\cdot Area(T)} \begin{pmatrix}s_{j}-s_{i} & s_{k}-s_{i}\\
 t_{j}-t_{i} & t_{k}-t_{i}\end{pmatrix} \begin{pmatrix}h_{k}-h_{i} & g_{i}-g_{k}\\
 h_{i}-h_{j} & g_{j}-g_{i} \end{pmatrix} \\
 & =  \begin{pmatrix}A_{T}^{i}s_{i}+A_{T}^{j}s_{j}+A_{T}^{k}s_{k} & B_{T}^{i}s_{i}+B_{T}^{j}s_{j}+B_{T}^{k}s_{k}\\
 A_{T}^{i}t_{i}+A_{T}^{j}t_{j}+A_{T}^{k}t_{k} & B_{T}^{i}t_{i}+B_{T}^{j}t_{j}+B_{T}^{k}t_{k} \end{pmatrix}.
 \end{align*}
\noindent where 
\begin{align*}
    &A_{T}^{i}=\dfrac{\left(h_{j}-h_{k}\right)}{2 Area(T)};\quad A_{T}^{j}=\dfrac{\left(h_{k}-h_{i}\right)}{2 Area(T)};\quad A_{T}^{k}=\dfrac{\left(h_{i}-h_{j}\right)}{2 Area(T)}; \\
    &B_{T}^{i}=\dfrac{\left(g_{k}-g_{j}\right)}{2 Area(T)};\quad B_{T}^{j}=\dfrac{\left(g_{i}-g_{k}\right)}{2 Area(T)};\quad B_{T}^{k}=\dfrac{\left(g_{j}-g_{i}\right)}{2 Area(T)};
\end{align*}

\begin{figure*}[t]
\centering
\begin{subfigure}[t]{0.16\textwidth}
\includegraphics[width=\textwidth]{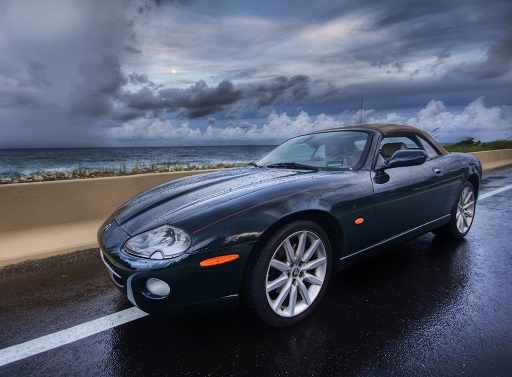}
\caption{}
\end{subfigure}
\begin{subfigure}[t]{0.16\textwidth}
\includegraphics[width=\textwidth]{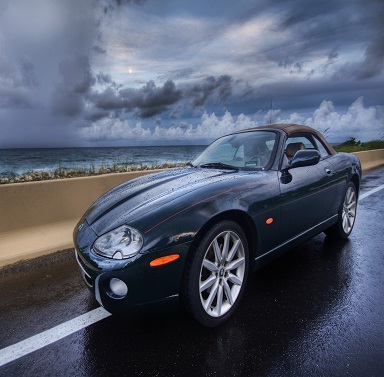}
\caption{}
\end{subfigure} 
\begin{subfigure}[t]{0.16\textwidth}
\includegraphics[width=\textwidth]{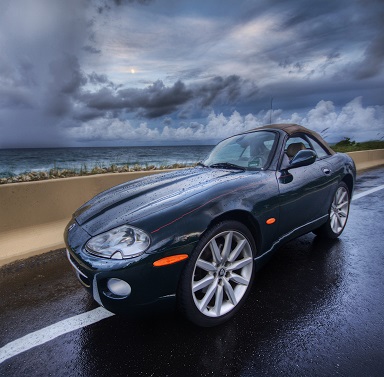}
\caption{}
\end{subfigure} 
\begin{subfigure}[t]{0.16\textwidth}
\includegraphics[width=\textwidth]{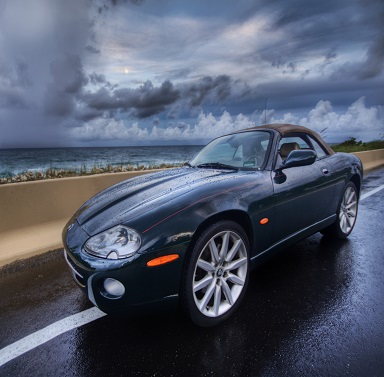}
\caption{}
\end{subfigure}
\begin{subfigure}[t]{0.16\textwidth}
\includegraphics[width=\textwidth]{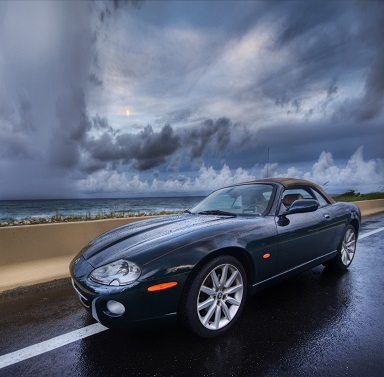}
\caption{}
\end{subfigure}
\begin{subfigure}[t]{0.16\textwidth}
\includegraphics[width=\textwidth]{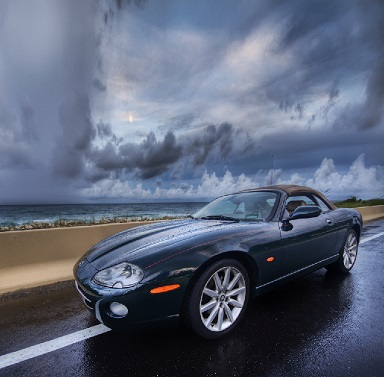}
\caption{}
\end{subfigure}
\caption{Comparsion of different methods to resize a car image to 75\% of the original width. (a) Original image. (b) Simple scaling. (c) Seam-Carving (SC). (d) Nonhomogeneous warping (Warp). (e) Scale-and-Stretch (SNS). (f) Proposed method.}
\label{fig:car}
\end{figure*}

The above gives a discretization of the gradient operator $D$. The discretization of the divergence operator is slightly more complicated. Here we want to take the divergence of a vector field defined on the faces, namely $(−d,c)_T$, and the divergence is then a function of the vertices (roughly speaking, while gradient is applied on the graph, the divergence is applied on the dual graph). Since divergence measures the net flux across the boundary normalized by area, we define the divergence of any vector field $(X_1,X_2)_T$ on faces to be:
\begin{equation}
\begin{split}
\text{Div}(X_{1},X_{2})(v_{i})=\sum_{T\in N_{i}}&Area(T) \cdot A_{T}^{i} X_{1}(T)\\
&+Area(T) \cdot B_{T}^{i}X_{2}(T).
\end{split}
\end{equation}
Here $N_i$ denote the set of faces which contain the vertex indexed with $i$. This is a right definition, since it is easy to check that for each $v_i$
\begin{equation}
\begin{split}
 & \text{Div}(-d,c)(v_{i}) \\ 
  =  &\sum_{T\in N_{i}}-Area(T) \cdot A_{T}^{i}d_{T}+Area(T) \cdot B_{T}^{i}c_{T}\\ 
  = & \sum_{T\in N_{1}}- Area(T) \cdot A_{T}^{i}\left( B_{T}^{i}t_{i}+ B_{T}^{j}t_{j}+B_{T}^{k}t_{k}\right)\\
 &+  Area(T) \cdot B_{T}^{i}\left(A_{T}^{i}t_{i}+ A_{T}^{j}t_{j}+A_{T}^{k}t_{k}\right)\\ 
  = &\ 0,
\end{split}
\end{equation}

\noindent where $c$ and $d$ are functions defined on each face such that $c(T) = c_T$ and $d(T) = d_T$. And similarly, ${\text{Div}(-b,a)(v_{i})=0}$, where $a$ and $b$ are functions defined on each face such that $a(T) = a_T$ and $b(T) = b_T$.

With the formulation of discrete gradient $D$ and discrete divergence Div operators, the elliptic PDE $\nabla \cdot  \bigg(A
\begin{pmatrix}
u_x \\
u_y
\end{pmatrix} \bigg)=0$ can be discretized as:
\begin{equation}\label{discreteu}
\begin{split}
\mathcal{L}(u)(v_i) = &\text{Div}\left\{ A D u\right\}(v_i)\\
=&\text{Div}\left\{ A \begin{bmatrix}A_{T}^{i}s_{i}+A_{T}^{j}s_{j}+A_{T}^{k}s_{k}\\ B_{T}^{i}s_{i}+B_{T}^{j}s_{j}+B_{T}^{k}s_{k} \end{bmatrix}\right\} =0
\end{split}
\end{equation}

Note that the above linear system must be satisfied for interior vertices. Let $\mathcal{V}_{int}$ and $\mathcal{V}_{b}$ be the index sets of the interior and boundary vertices respectively. Together with the boundary constraint, the linear system (\ref{discreteu}) can be written as $L^u \overrightarrow{s} = \overrightarrow{b^u}$, where $\overrightarrow{s} = (s_1,s_2,...,s_n)$ ($n$ is the number of vertices) and
\begin{equation}
    \begin{cases}
        L^u(i,j)=0 & i \neq j \\
        L^u(i,i)=1
    \end{cases}
\end{equation}
\noindent for all $i\in \mathcal{V}_{b}$. $\overrightarrow{b^u}$ is determined by the boundary constraints:
\begin{equation}
     \begin{cases}
        b^u_i=0 & \text{if }i\in \mathcal{V}_{b} \text{ and } v_i \text{ is on the left boundary,} \\
        b^u_i=m' & \text{if } i\in \mathcal{V}_{b} \text{ and } v_i \text{ is on the right boundary,}\\
        b^u_i=0 & \text{otherwise.}\\
    \end{cases}
\end{equation}

Similarly, the elliptic PDE 
$\nabla \cdot  \bigg(A \begin{pmatrix}
v_x \\
v_y
\end{pmatrix} \bigg)=0$ with the boundary constraints can be discretized as: $L^v \overrightarrow{t} = \overrightarrow{b^v}$, $\overrightarrow{t} = (t_1,t_2,...,t_n)$ and
\begin{equation}
    \begin{cases}
        L^v(i,j)=0 & i \neq j \\
        L^v(i,i)=1
    \end{cases}
\end{equation}
\noindent for all $i\in \mathcal{V}_{b}$. $\overrightarrow{b^v}$ is determined by the boundary constraints:
\begin{equation}
     \begin{cases}
        b^v_i=0 & \text{if }i\in \mathcal{V}_{b} \text{ and } v_i \text{ is on the bottom boundary,} \\
        b^v_i=n' & \text{if }i\in \mathcal{V}_{b} \text{ and } v_i \text{ is on the top boundary,} \\
        b^v_i=0 & \text{otherwise.}
    \end{cases}
\end{equation}

As a result, the elliptic PDE (\ref{equation:elliptic PDE}) can be discretized as the following linear system:
\begin{equation}\label{generalizedlinear}
    L\begin{pmatrix}
    \overrightarrow{s} \\
    \overrightarrow{t}
    \end{pmatrix}:=\begin{pmatrix}
    L^u & \overrightarrow{0} \\
    \overrightarrow{0} & L^v
    \end{pmatrix}
    \begin{pmatrix}
    \overrightarrow{s} \\
    \overrightarrow{t}
    \end{pmatrix}
    =\begin{pmatrix}
    \overrightarrow{b^u} \\
    \overrightarrow{b^v}
    \end{pmatrix}
\end{equation}

\begin{figure*}[t]
\centering
\begin{subfigure}[t]{0.16\textwidth}
\includegraphics[width=\textwidth]{tajmahal}
\caption{}
\end{subfigure}
\begin{subfigure}[t]{0.16\textwidth}
\includegraphics[width=\textwidth]{tajmahal_landmark}
\caption{}
\end{subfigure} 
\begin{subfigure}[t]{0.16\textwidth}
\includegraphics[width=\textwidth]{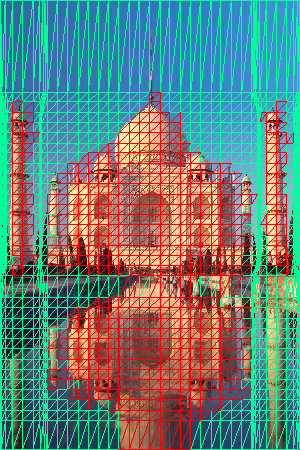}
\caption{}
\end{subfigure}
\begin{subfigure}[t]{0.16\textwidth}
\includegraphics[width=\textwidth]{tajmahal_weak}
\caption{}
\end{subfigure}
\begin{subfigure}[t]{0.16\textwidth}
\includegraphics[width=\textwidth]{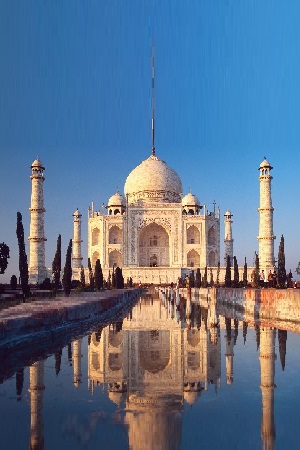}
\caption{}
\end{subfigure}
\begin{subfigure}[t]{0.16\textwidth}
\includegraphics[width=\textwidth]{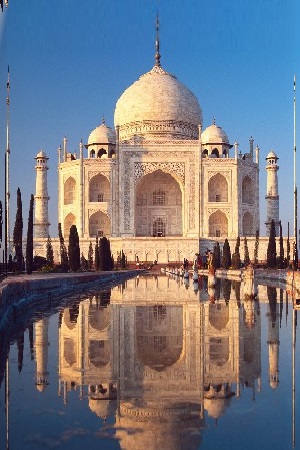}
\caption{}
\end{subfigure}
\caption{Proposed method with Choice 2. (a) Original image. (b) Image with labeled important objects. (c) Resized image to 50\% of the original width using Choice 2. The deformed mesh is also shown. (d) Resized image using Choice 2. (e) Resized image with Choice 1. (f) Resized image with Choice 3.}
\label{fig:weak mu}
\end{figure*}

Besides, the deformation type constraints in the object regions and line structures should also be enforced. For simplicity, we consider the situation when there are only one object region and one line structure. The case when there are more than one objects and lines can be tackled similarly. Denote the meshes of the object regions $O$ and the line structures $\mathit{l}$ by $\mathcal{K}(O)$ and $\mathcal{K}(l)$. In the discrete case, we impose that
\begin{equation}
\begin{split}
f(v_i) &= r^O v_i + \overrightarrow{t}^O \text{ for }v_i \in \mathcal{K}(O)\\
f(v_i) &= r^l v_i + \overrightarrow{t}^l \text{ for }v_i \in \mathcal{K}(l)\\
\end{split}
\end{equation}

We then require that 
\begin{equation}
\begin{split}
    \mathcal{L} f (v_i)&=r^O \mathcal{L}(x)(v_i) + \overrightarrow{t}^O  = \overrightarrow{0} \text{ for }v_i\in \mathcal{K}(O)\\
\mathcal{L}f (v_i) &=\vec{r}^l \cdot \mathcal{L}(x)(v_i) + \overrightarrow{t}^l = \overrightarrow{0}\text{ for }v_i\in \mathcal{K}(l)
\end{split}
\end{equation}

By rearranging $\begin{pmatrix}
    \overrightarrow{s} \\
    \overrightarrow{t}
    \end{pmatrix}$, the linear system with the deformation type constraints can be written as:
    
    \begin{equation}
    \begin{pmatrix}
    \tilde{L}^u & \overrightarrow{0} & \overrightarrow{w^x_o} & \overrightarrow{s_O} & \overrightarrow{0} & \overrightarrow{w^x_l}  & \overrightarrow{0} & \overrightarrow{s_l} & \overrightarrow{0} \\
    \overrightarrow{0} & \tilde{L}^v & \overrightarrow{w^y_o} & \overrightarrow{0} & \overrightarrow{s_O} & \overrightarrow{0} & \overrightarrow{w^y_l} & \overrightarrow{0} & \overrightarrow{s_l} 
    \end{pmatrix}
    \begin{pmatrix}
    \overrightarrow{s'} \\
    \overrightarrow{t'} \\
    r_o \\
    t^x_o \\
    t^y_o \\
    r^x_l \\
    r^y_l \\
    t^x_l \\
    t^y_l 
    \end{pmatrix}
    =\begin{pmatrix}
    \overrightarrow{\tilde{b}^u} \\
    \overrightarrow{\tilde{b}^v}
    \end{pmatrix}
\end{equation}
\noindent where 
$$\overrightarrow{w_O^x}=\sum\limits_{ v_i \in \mathcal{K}(O)} g_i
    \begin{pmatrix}
    | \\
    \overrightarrow{L_i} \\
    |
    \end{pmatrix},\ \overrightarrow{w_O^y}=\sum\limits_{ v_i \in \mathcal{K}(O)} h_i
    \begin{pmatrix}
    | \\
    \overrightarrow{L_i} \\
    |
    \end{pmatrix},$$
    $$\overrightarrow{w_l^x}=\sum\limits_{ v_i \in \mathcal{K}(l)} g_i
    \begin{pmatrix}
    | \\
    \overrightarrow{L_i} \\
    |
    \end{pmatrix},\ \overrightarrow{w_l^y}=\sum\limits_{ v_i \in \mathcal{K}(l)} h_i
    \begin{pmatrix}
    | \\
    \overrightarrow{L_i} \\
    |
    \end{pmatrix},$$
        $$\overrightarrow{s_O}=\sum\limits_{ v_i \in \mathcal{K}(O)} \begin{pmatrix}
    | \\
    \overrightarrow{L_i} \\
    |
    \end{pmatrix} \text{ and } \overrightarrow{s_l}=\sum\limits_{ v_i \in \mathcal{K}(l)} \begin{pmatrix}
    | \\
    \overrightarrow{L_i} \\
    |
    \end{pmatrix}.$$

\begin{figure*}[t]
\centering
\begin{subfigure}[t]{0.48\textwidth}
\includegraphics[width=\textwidth]{tajmahal}
\caption{}
\end{subfigure}
\begin{subfigure}[t]{0.24\textwidth}
\includegraphics[width=\textwidth]{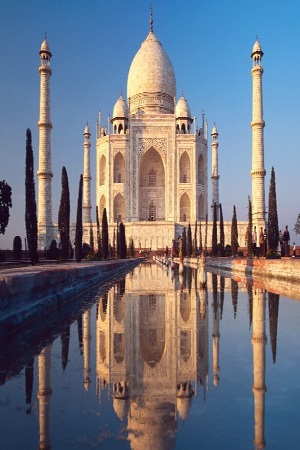}
\caption{}
\end{subfigure} 
\begin{subfigure}[t]{0.24\textwidth}
\includegraphics[width=\textwidth]{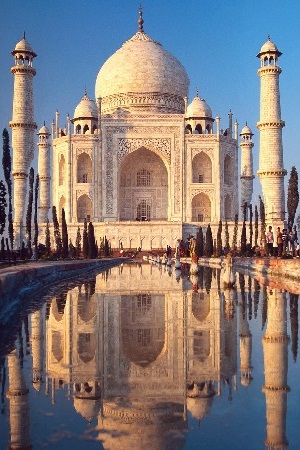}
\caption{}
\end{subfigure} \\
\begin{subfigure}[t]{0.24\textwidth}
\includegraphics[width=\textwidth]{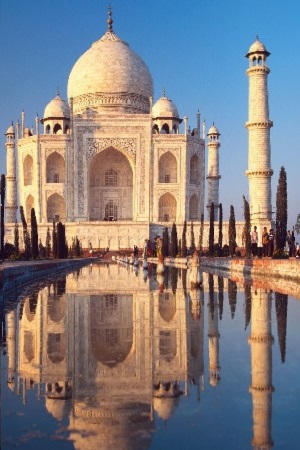}
\caption{}
\end{subfigure}
\begin{subfigure}[t]{0.24\textwidth}
\includegraphics[width=\textwidth]{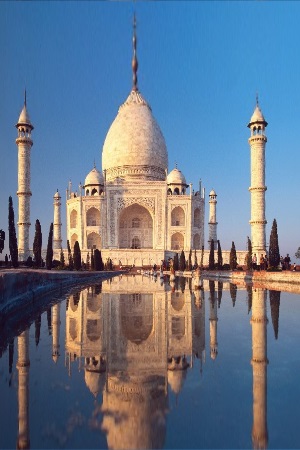}
\caption{}
\end{subfigure}
\begin{subfigure}[t]{0.24\textwidth}
\includegraphics[width=\textwidth]{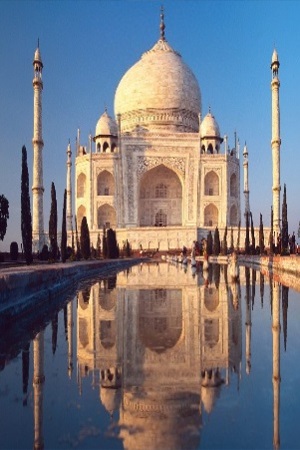}
\caption{}
\end{subfigure}
\begin{subfigure}[t]{0.24\textwidth}
\includegraphics[width=\textwidth]{tajmahal_weak}
\caption{}
\end{subfigure}
\caption{Comparsion of different methods to resize a Taj Maha image to 50\% of the original width. (a) Original image. (b) Simple scaling. (c) Seam-Carving (SC). (d) Nonhomogeneous warping (Warp). (e) Scale-and-Stretch (SNS). (f) Energy-based deformation (LG). (g) Proposed method.}
\label{fig:tajmahal}
\end{figure*}

\noindent Here, $\overrightarrow{s'}$ and $\overrightarrow{t'}$ are respectively the $u$ and $v$ coordinates of the vertices in the homogeneous region. $\overrightarrow{\tilde{b}^u}$ and $\overrightarrow{\tilde{b}^v}$ are obtained by permutations of  $\overrightarrow{b^u}$ and $\overrightarrow{b^v}$ respectively.

The chessboard constraints can also be imposed similarly. Denote the meshes of $\mathcal{M}_h$ and $\mathcal{M}_v$ by $\mathcal{K}(\mathcal{M}_h)$ and $\mathcal{K}(\mathcal{M}_v)$ respectively. Again, we assume that there is only one object in the image for illustration purpose. Note that we require that
\begin{equation}
\begin{split}
    \mathcal{L} f (v_i)&=r \mathcal{L}(x)(v_i) + \begin{pmatrix}
    c^x \\
    0
    \end{pmatrix}  = \overrightarrow{0} \text{ for }v_i\in \mathcal{K}(\mathcal{M}_h)\\
\mathcal{L}f (v_i) &=r \mathcal{L}(x)(v_i) + \begin{pmatrix}
    0 \\
    c^y
    \end{pmatrix} = \overrightarrow{0}\text{ for }v_i\in \mathcal{K}(\mathcal{M}_v)
\end{split}
\end{equation}

By rearranging $\begin{pmatrix}
    \overrightarrow{s} \\
    \overrightarrow{t}
    \end{pmatrix}$, the linear system with the chessboard constraints can be written as:

\begin{equation}
    \begin{pmatrix}
    \tilde{L}^u & \overrightarrow{0} & \overrightarrow{w^x} & \overrightarrow{s^{x}} & \overrightarrow{0}  \\
    \overrightarrow{0} & \tilde{L}^v & \overrightarrow{w^y} & \overrightarrow{0} & \overrightarrow{s^{y}}  
    \end{pmatrix}
    \begin{pmatrix}
    \overrightarrow{s'} \\
    \overrightarrow{t'} \\
    r \\
    c_i^x \\
    c_i^y 
    \end{pmatrix}
    =\begin{pmatrix}
    \overrightarrow{\tilde{b}^u} \\
    \overrightarrow{\tilde{b}^v}
    \end{pmatrix}
\end{equation}
    
\noindent where
    $$\overrightarrow{w^x}=\sum\limits_{ v_i \in \mathcal{K}(\mathcal{M}_h)} g_i  \begin{pmatrix}
    | \\
    \overrightarrow{L_i} \\
    |
    \end{pmatrix};\ \overrightarrow{w^y}=\sum\limits_{ v_i \in \mathcal{K}(\mathcal{M}_v)} h_i  \begin{pmatrix}
    | \\
    \overrightarrow{L_i} \\
    |
    \end{pmatrix};$$
$$
    \overrightarrow{s^x}=\sum\limits_{ v_i \in\mathcal{K}(\mathcal{M}_h)} \begin{pmatrix}
    | \\
    \overrightarrow{L_i} \\
    |
    \end{pmatrix} \text{ and } \overrightarrow{s^y}=\sum\limits_{ v_i \in\mathcal{K}(\mathcal{M}_v)} \begin{pmatrix}
    | \\
    \overrightarrow{L_i} \\
    |
    \end{pmatrix}.
$$

\begin{figure*}[t]
\centering
\begin{subfigure}[t]{0.32\textwidth}
\includegraphics[width=\textwidth]{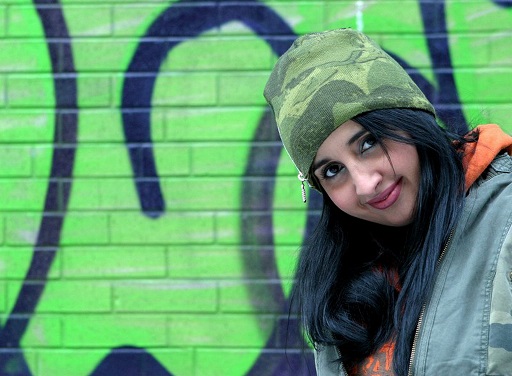}
\caption{}
\end{subfigure}
\begin{subfigure}[t]{0.24\textwidth}
\includegraphics[width=\textwidth]{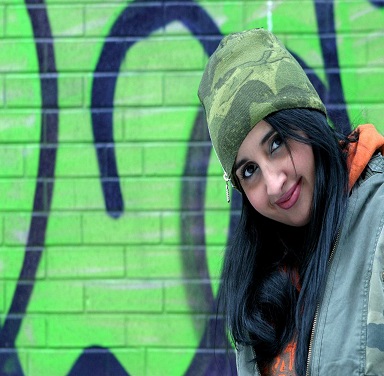}
\caption{}
\end{subfigure} 
\begin{subfigure}[t]{0.24\textwidth}
\includegraphics[width=\textwidth]{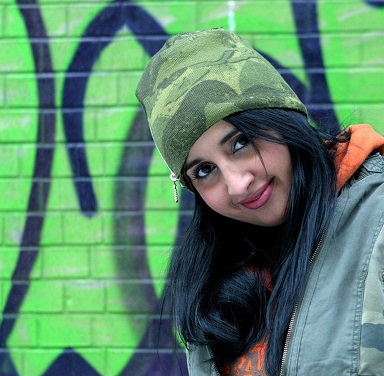}
\caption{}
\end{subfigure} \\
\begin{subfigure}[t]{0.24\textwidth}
\includegraphics[width=\textwidth]{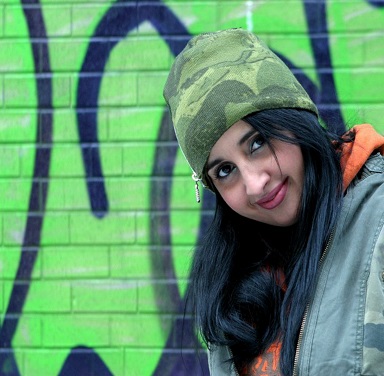}
\caption{}
\end{subfigure}
\begin{subfigure}[t]{0.24\textwidth}
\includegraphics[width=\textwidth]{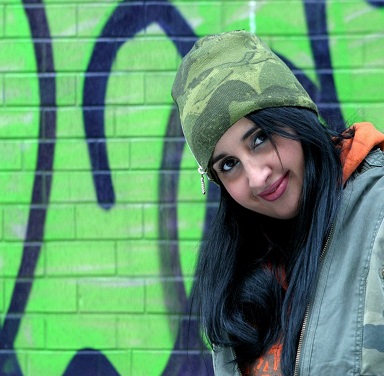}
\caption{}
\end{subfigure}
\begin{subfigure}[t]{0.24\textwidth}
\includegraphics[width=\textwidth]{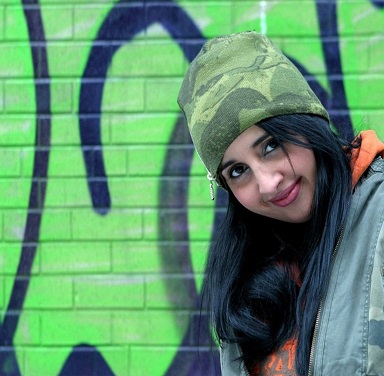}
\caption{}
\end{subfigure}
\begin{subfigure}[t]{0.24\textwidth}
\includegraphics[width=\textwidth]{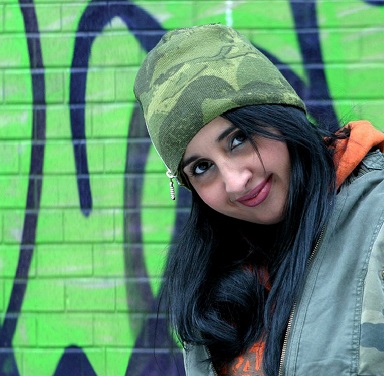}
\caption{}
\end{subfigure}
\caption{Comparsion of different methods to resize a portrait image to 75\% of the original width. (a) Original image. (b) Simple scaling. (c) Seam-Carving (SC). (d) Nonhomogeneous warping (Warp). (e) Scale-and-Stretch (SNS). (f) Energy-based deformation (LG). (g) Proposed method.}
\label{fig:Fatem}
\end{figure*}

\begin{figure*}[t]
\centering
\begin{subfigure}[t]{0.16\textwidth}
\includegraphics[width=\textwidth]{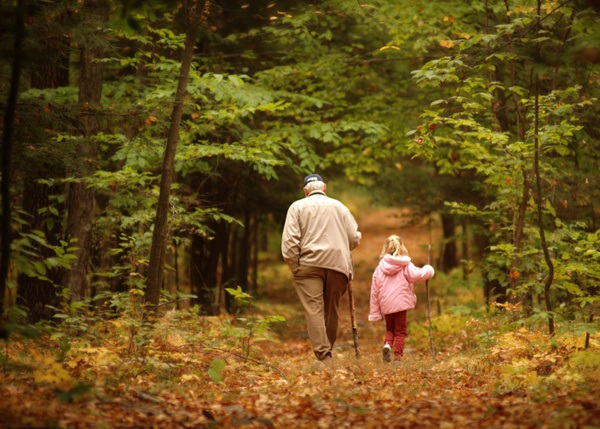}
\caption{}
\end{subfigure}
\begin{subfigure}[t]{0.16\textwidth}
\includegraphics[width=\textwidth]{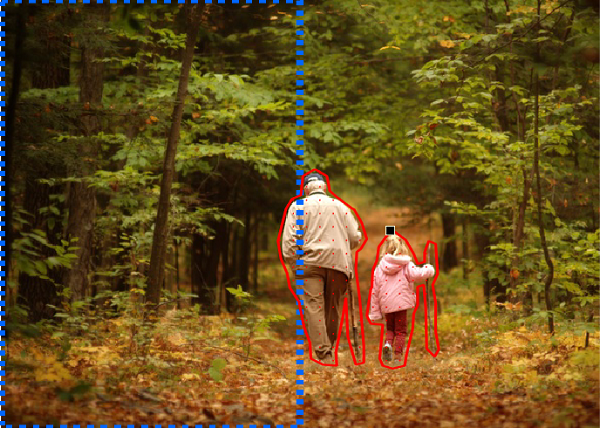}
\caption{}
\end{subfigure} 
\begin{subfigure}[t]{0.16\textwidth}
\includegraphics[width=\textwidth]{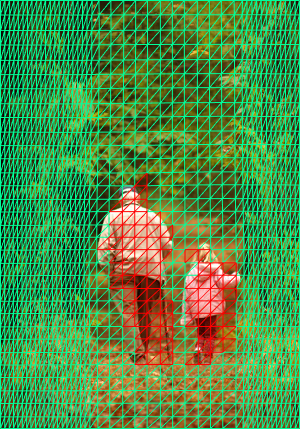}
\caption{}
\end{subfigure}
\begin{subfigure}[t]{0.16\textwidth}
\includegraphics[width=\textwidth]{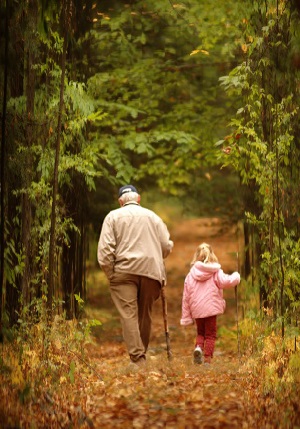}
\caption{}
\end{subfigure}
\begin{subfigure}[t]{0.16\textwidth}
\includegraphics[width=\textwidth]{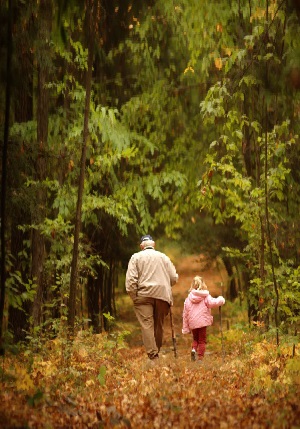}
\caption{}
\end{subfigure}
\begin{subfigure}[t]{0.16\textwidth}
\includegraphics[width=\textwidth]{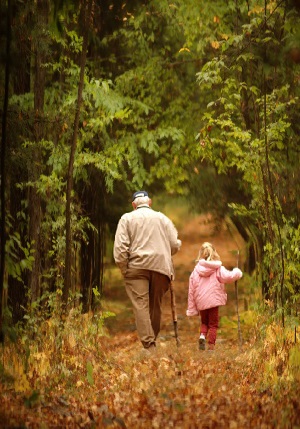}
\caption{}
\end{subfigure}
\caption{Proposed method with Choice 3. (a) Original image. (b) Image with labeled important objects labeled. (c) Retargeted image with a mesh. (d) Resized image to 50\% of the original width with Choice 3. (e) Resized image with Choice 1. (f) Resized image with Choice 2.}
\label{fig:strong mu}
\end{figure*}

\section{Results and discussion} \label{experiment}
In this section, experimental results will be illustrated in details. The retargeted images obtained by our proposed algorithm with different feature conditions and constraints are compared with some representative methods such as Seam Carving (SC)\cite{Rubinstein2008}, Nonhomogenous warping (Warp)\cite{Wolf2007}, Scale-and-Stretch (SNS)\cite{Wang2008} and Energy-based deformation (LG)\cite{Karni2009}. The images used in our experiments are obtained from \cite{Rubinstein2010}. The effectiveness of our proposed method will be assessed by its ability to handle images with different complexity. In each example, the resized image by simple scaling is also shown to demonstrate the importance of developing a better image retargeting algorithm. The proposed algorithm is implemented in Matlab\footnote{The proposed algorithm can also be implemented in mobile devices: \url{https://github.com/Edward-Yung/Mu-Retargeter-1.0}}. All the experiments are executed on an Intel Core i7 3.4GHz computer. On average, the computation time to obtain the warping map by our proposed method is about a second with $1500$ vertices on the regular triangular mesh.

\subsection{Even distribution of distortions} \label{exp:Mono mu}

In Figure \ref{fig:mono mu}, the retargeted results of a $615 \times 461$ image to a $461 \times 461$ image are shown. The penguins are the important objects. Since there are several objects and the position of the objects are scattered, Choice 1 is applied to prescribe the BR. It can be observed that the geometric structure of the penguins is well-preserved while the distortion of the background is insignificant. On the other hand, the retargeted results of using Choice 2 and Choice 3 to prescribe the BR are shown in Figure \ref{fig:mono mu} (e) and \ref{fig:mono mu} (f) respectively. Note that undesirable artifacts can be found in the rock at the top right corner since Choice 2 and 3 tolerate more distortion in the background. It suggests that Choice 1 should be applied if the object regions occupy a large portion of the image.

Besides, we also compare our method with other existing approaches. The retargeting results of the penguin image by different approaches are as shown in Figure \ref{fig:penguin}. With the SC method, the penguins on the middle left cannot be preserved and some parts are truncated. With the Warp method and SNS method, the shape of the second penguin from the left is shrunk. Using our proposed method, the shapes of all the penguins are well-preserved. Figure \ref{fig:car} shows the retargeting results of the car image. With the SC method, the car is severely distorted and the white line on the road is bent. With the Warp method, the car is shrunk and the white line is also bent. With the SNS method, the line structures are kept and the car is slightly deformed. Using our proposed method, the shape of the car is kept, the line structures are well-preserved and the texture details in the background are also preserved.

\begin{figure*}[t]
\centering
\begin{subfigure}[t]{0.48\textwidth}
\includegraphics[width=\textwidth]{family}
\caption{}
\end{subfigure}
\begin{subfigure}[t]{0.24\textwidth}
\includegraphics[width=\textwidth]{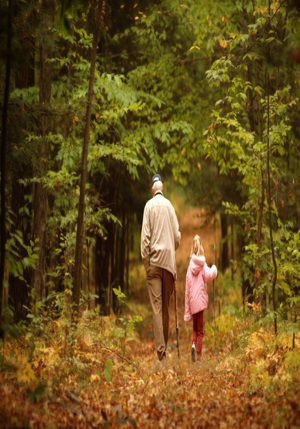}
\caption{}
\end{subfigure} 
\begin{subfigure}[t]{0.24\textwidth}
\includegraphics[width=\textwidth]{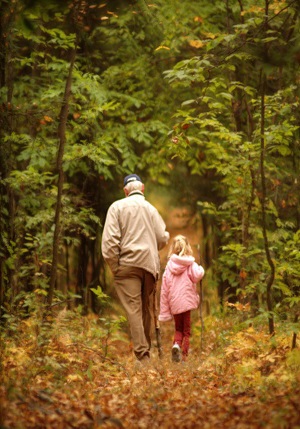}
\caption{}
\end{subfigure} \\
\begin{subfigure}[t]{0.24\textwidth}
\includegraphics[width=\textwidth]{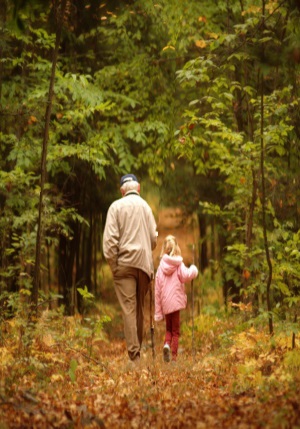}
\caption{}
\end{subfigure}
\begin{subfigure}[t]{0.24\textwidth}
\includegraphics[width=\textwidth]{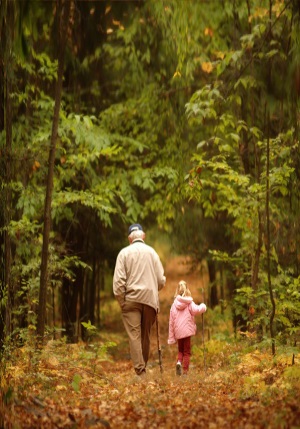}
\caption{}
\end{subfigure}
\begin{subfigure}[t]{0.24\textwidth}
\includegraphics[width=\textwidth]{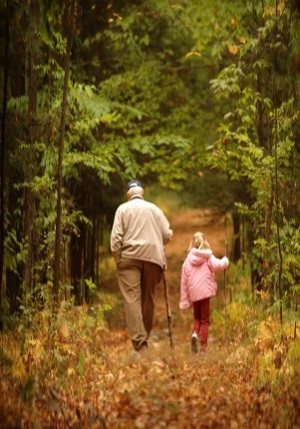}
\caption{}
\end{subfigure}
\begin{subfigure}[t]{0.24\textwidth}
\includegraphics[width=\textwidth]{family_strong_line}
\caption{}
\end{subfigure}
\caption{Comparsion of different methods to resize a two-person image to 50\% of the original width. (a) Original image. (b) Simple scaling. (c) Seam-Carving (SC). (d) Nonhomogeneous warping (Warp). (e) Scale-and-Stretch (SNS). (f) Energy-based deformation (LG). (g) Proposed method.}
\label{fig:family}
\end{figure*}

\subsection{Weakly uneven distribution of distortions} \label{exp:Weak mu}
In Figure \ref{fig:weak mu}, the retargeting results of a $600 \times 450$ image to a $300 \times 450$ image are shown. The Taj Maha is the important object. We set a large resizing ratio, which is $50\%$. In this example, Choice 2 is applied to prescribe the BR, since the background is considered as unimportant and can tolerate larger distortions. The resized image is shown in Figure \ref{fig:weak mu}(d). It can be observed that the size of Taj Maha can be well maintained. Note that the mesh between the important regions is still foldover-free, despite the large resizing ratio (see Figure \ref{fig:weak mu}(c)). Figure \ref{fig:weak mu}(e) shows the retargeted result using Choice 1. The size of Taj Maha becomes noticeably smaller since Choice 1 distributes the distortion evenly over the whole image. Figure \ref{fig:weak mu}(f) shows the retargeted result using Choice 3. The result is generally not satisfactory, although the size of Taj Maha is better maintained. Some details in the image are missing. For example, the pillars on the two sides are removed.

In Figure \ref{fig:tajmahal}, the results obtained by various approaches are shown. With the SC method, some parts of the pillars are removed. With the Warp method, the image is slightly distorted and one pillar is gone. With the SNS method and LG method, the object is severely shrunk. Using our proposed method, the geometric structure of Taj Maha is well-preserved. In Figure \ref{fig:Fatem}, we compare various image retargeting methods on a portrait image. With the Warp method, the shape of the woman is deformed. With the SNS method, the shape of the woman is deformed and the shape of her hat is distorted. With the LG method, the shape of the woman is shrunk. Also, the line structure behind her is shrunk. With SC method and our proposed method, both can get a similar result in the important region of the image. However, the background texture on the left is deformed with the SC method while our proposed method can better keep the texture.

\begin{figure*}[t]
\centering
\begin{subfigure}[t]{0.32\textwidth}
\includegraphics[width=\textwidth]{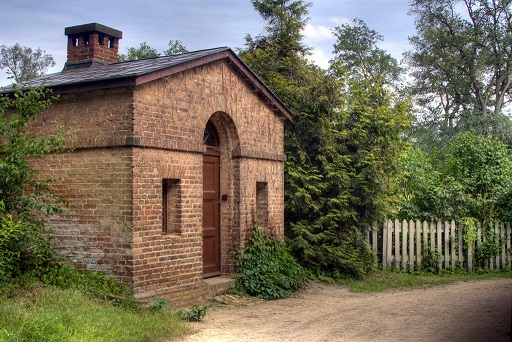}
\caption{}
\end{subfigure}
\begin{subfigure}[t]{0.24\textwidth}
\includegraphics[width=\textwidth]{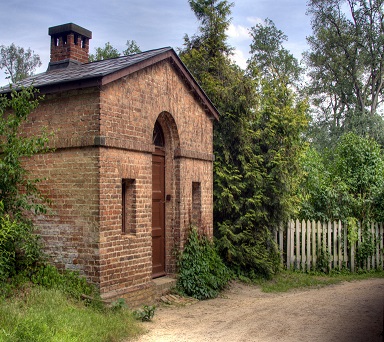}
\caption{}
\end{subfigure} 
\begin{subfigure}[t]{0.24\textwidth}
\includegraphics[width=\textwidth]{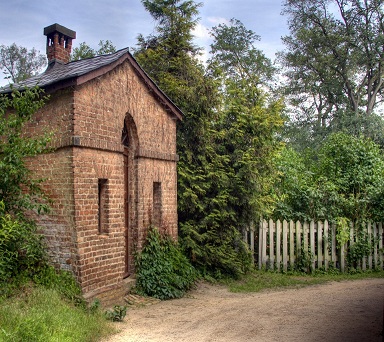}
\caption{}
\end{subfigure} \\
\begin{subfigure}[t]{0.24\textwidth}
\includegraphics[width=\textwidth]{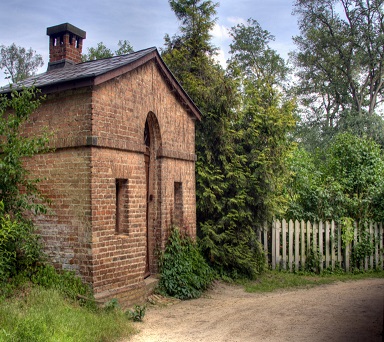}
\caption{}
\end{subfigure}
\begin{subfigure}[t]{0.24\textwidth}
\includegraphics[width=\textwidth]{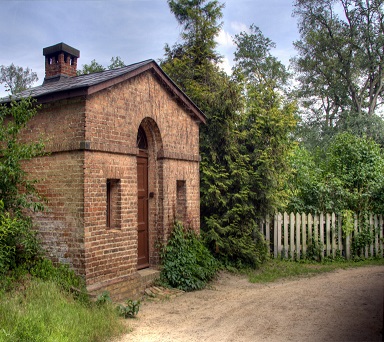}
\caption{}
\end{subfigure}
\begin{subfigure}[t]{0.24\textwidth}
\includegraphics[width=\textwidth]{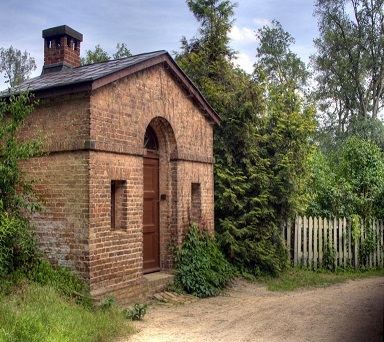}
\caption{}
\end{subfigure}
\begin{subfigure}[t]{0.24\textwidth}
\includegraphics[width=\textwidth]{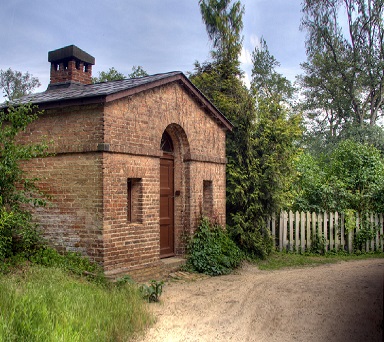}
\caption{}
\end{subfigure}
\caption{Comparsion of different methods to resize a brick house image to 75\% of the original width. (a) Original image. (b) Simple scaling. (c) Seam-Carving (SC). (d) Nonhomogeneous warping (Warp). (e) Scale-and-Stretch (SNS). (f) Energy-based deformation (LG). (g) Proposed method.}
\label{fig:brick_house}
\end{figure*}

\subsection{Strongly uneven distribution of distortions} \label{exp:Strong mu}

In Figure \ref{fig:strong mu}, the resized results of a $600 \times 429$ image to a $300 \times 429$ image are shown. The two persons (the old man and the girl) are the important objects in the image. Our goal is to preserve the geometry of the two persons and the textures around them as good as possible while tolerating some deformations of the background. In this example, Choice 2 is applied to prescribe the BR. The result is shown in Figure \ref{fig:strong mu}(d). Note that the shape of the two persons is well-preserved. Also, the texture in the background is only a little bit distorted. The sizes of the two persons are almost the same as the original image. With Choice 1, the sizes of the two persons become much smaller, as shown in Figure \ref{fig:strong mu} (e). On the other hand, the texture in the background is significantly distorted using Choice 2, as shown in Figure \ref{fig:strong mu} (f).

In Figure \ref{fig:family}, the results of our proposed methods are compared with other existing approaches. With the SC method, the two persons are significantly distorted. With the Warp method and LG method, the two persons are shrunk. With the SNS method, the two persons are shrunk and become smaller. Using our proposed method with Choice 1 to prescribe the BR, the shapes and sizes of the two persons are well-preserved. We have also compared our methods with other existing approaches on a brick house image. The results are shown in Figure \ref{fig:brick_house}. The brick house is landmarked and line constraint is also enforced on the fence. With the SC method, the brick house is severely distorted. The edges of the roof become zigzag and the chimney is deformed. With the Warp method, the brick house is shrunk and the door is removed. With the SNS method, the brick house is shrunk. On the other hand, with the LG method, the brick house is deformed and the fence is shrunk. Using our proposed method, the shape of the brick house is well-preserved and the line structure of the fence is also maintained.

\subsection{Chessboard constraint} \label{exp:Chessboard constraints}
\begin{figure*}[t]
\centering
\begin{subfigure}[t]{0.24\textwidth}
\includegraphics[width=\textwidth]{car}
\caption{}
\end{subfigure}
\begin{subfigure}[t]{0.18\textwidth}
\includegraphics[width=\textwidth]{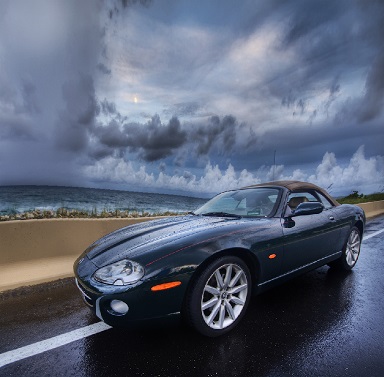}
\caption{}
\end{subfigure} 
\begin{subfigure}[t]{0.18\textwidth}
\includegraphics[width=\textwidth]{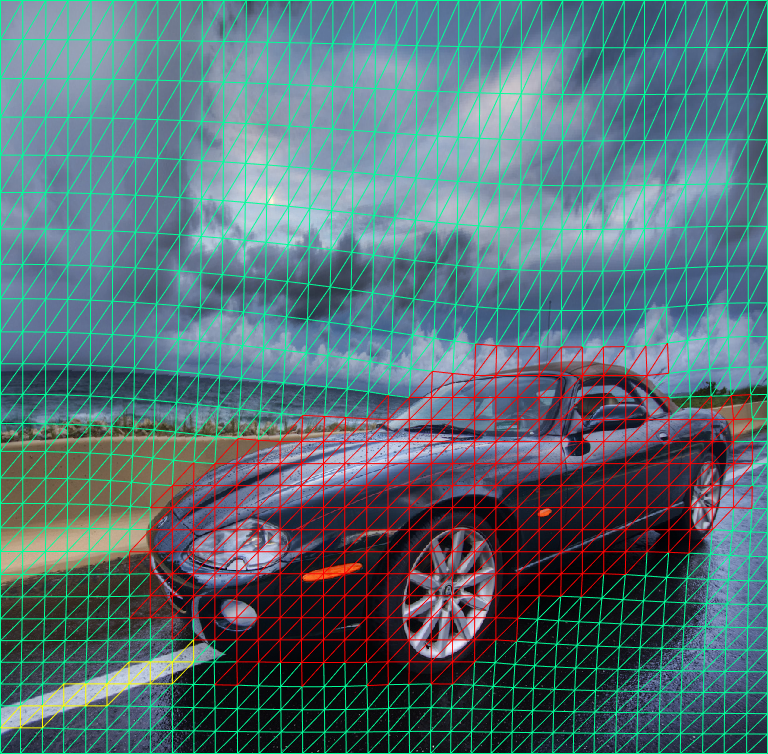}
\caption{}
\end{subfigure}
\begin{subfigure}[t]{0.18\textwidth}
\includegraphics[width=\textwidth]{car_mono_chsbd}
\caption{}
\end{subfigure}
\begin{subfigure}[t]{0.18\textwidth}
\includegraphics[width=\textwidth]{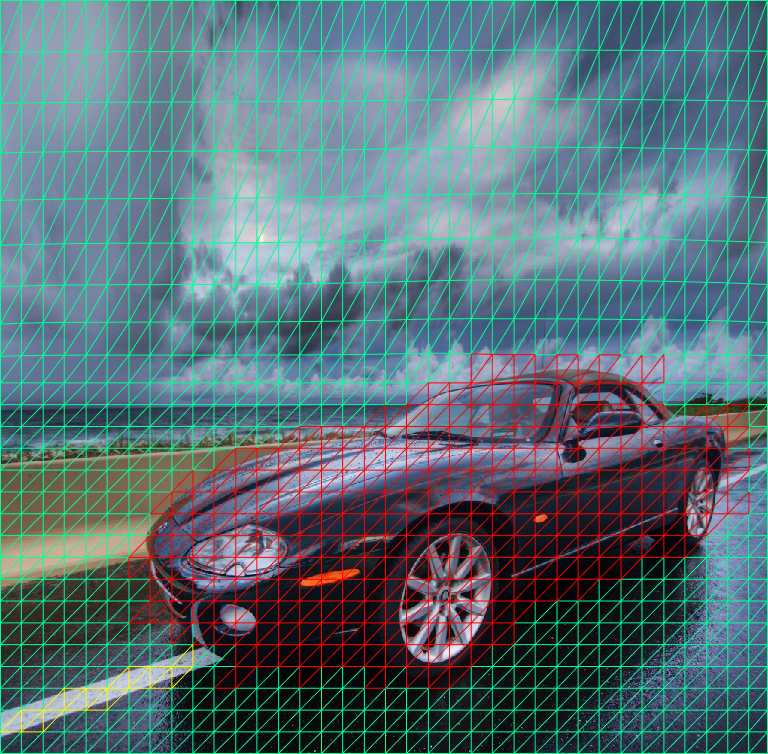}
\caption{}
\end{subfigure}
\caption{Retargeting result by our proposed method with the chessboard constraint. (a) Original image. (b) Resized image with Choice 1. (c) Deformed mesh on image (b). (d) Resized image with Choice 1 and chessboard constraint. (e) Deformed mesh on image (d).}
\label{fig:chessboard}
\end{figure*}

In some cases, the background in the image contains general line textures. If the image is just warped without any additional constraints, it may lead to unsatisfactory results. In Figure \ref{fig:chessboard}, the resized result using Choice 1 with and without the chessboard constraint are demonstrated. The image is retargeted from $1024 \times 754$ to a $768 \times 754$. Figure \ref{fig:chessboard}(b) shows the result without enforcing chessboard constraints. The curb near the car is originally straight but is distorted after resizing. By enforcing the chessboard constraints, the curb remains straight as shown in Figure \ref{fig:chessboard}(d).

\subsection{Extremal cases} \label{exp:Extreme cases}

\begin{figure*}[t]
\centering

\begin{subfigure}[t]{0.36\textwidth}
\includegraphics[width=\textwidth]{family}
\caption{}
\end{subfigure}
\begin{subfigure}[t]{0.119\textwidth}
\includegraphics[width=\textwidth]{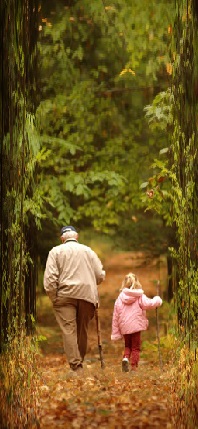}
\caption{}
\end{subfigure}
\begin{subfigure}[t]{0.345\textwidth}
\includegraphics[width=\textwidth]{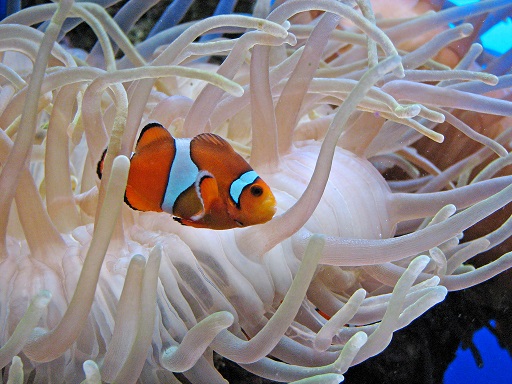}
\caption{}
\end{subfigure}
\begin{subfigure}[t]{0.1135\textwidth}
\includegraphics[width=\textwidth]{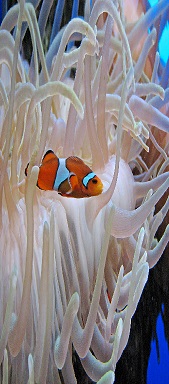}
\caption{}
\end{subfigure}

\caption{Image retargeting in the extremal case. (a) and (c) show two original images. (b) and (d) shows the resized image of(a) and (c) respectively.}
\label{fig:extreme}
\end{figure*}

In some extremal situations, the size of the important regions in the original image is even larger than the size of the resized image. Our proposed methods work effectively even in such a case. Note that the warping map obtained from the prescribed BR is bijective. Hence, a foldover-free warping of the image is always ensured even in such an extremal situation. Figure \ref{fig:extreme} shows two retargeting results in the extremal cases. The height of the image is kept while the width is reduced to one-fourth of the original image. Figure \ref{fig:extreme}(b) and Figure \ref{fig:extreme}(d) show the retargeted results of the two images. It can be observed that the geometric structures of the important objects are still well-preserved in the extremal cases.

\section{Conclusion and Future Work} \label{conclusion}
In this paper, we have proposed a simple and effective image retargeting method based on Beltrami representation. Our proposed algorithm using Beltrami representation for image resizing can preserve the geometric structures of important regions including objects and line structures while maintaining the background as good as possible. By reconstructing the warping map from the Beltrami representation, we can easily control the distortion in the unimportant region while preserving the shapes of important objects. Bijective warping map is guaranteed so as to avoid any undesirable foldover artifacts. Our proposed method can also deal with large resizing ratio in the extremal cases and give satisfactory results. No optimization procedure is needed and no hyper-parameter is required to be fine-tuned throughout the retargeting process. Hence, a fast computation can be achieved. In the future, we will explore the possibility to extend our method to deal with video retargeting problems. 

\vspace{-4mm}

\section*{Acknowledgments}
L.M. Lui is supported by RGC GRF (Project ID: 402413).


\vspace{-15mm}

\begin{IEEEbiography}[{\includegraphics[width=1in,height=1.25in,clip,keepaspectratio]{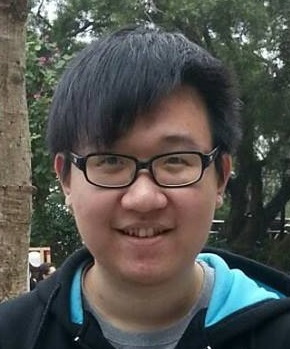}}]{Chun Pong Lau} received the BSc degree in mathematics at The Chinese University of Hong Kong (CUHK) in 2016. He is currently working towards the MPhil degree in mathematics at CUHK. His research interests include image and video processing (restoration and enhancement), computer vision and machine learning.
\end{IEEEbiography}

\vspace{-15mm}

\begin{IEEEbiography}[{\includegraphics[width=1in,height=1.25in,clip,keepaspectratio]{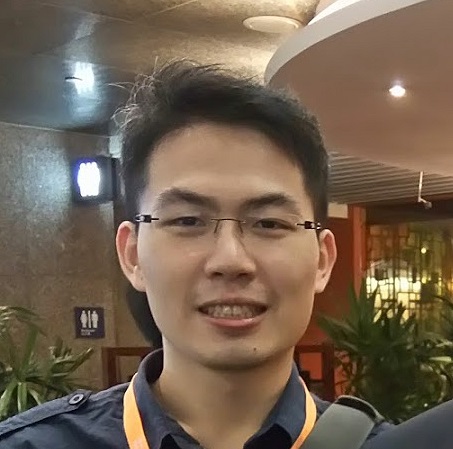}}]{Chun Pang Yung} received the BSc degree in mathematics at The Chinese University of Hong Kong (CUHK) in 2015. He is currently working as a software engineer. His research interests include image processing and computer vision.
\end{IEEEbiography}

\vspace{-15mm}

\begin{IEEEbiography}[{\includegraphics[width=1in,height=1.25in,clip,keepaspectratio]{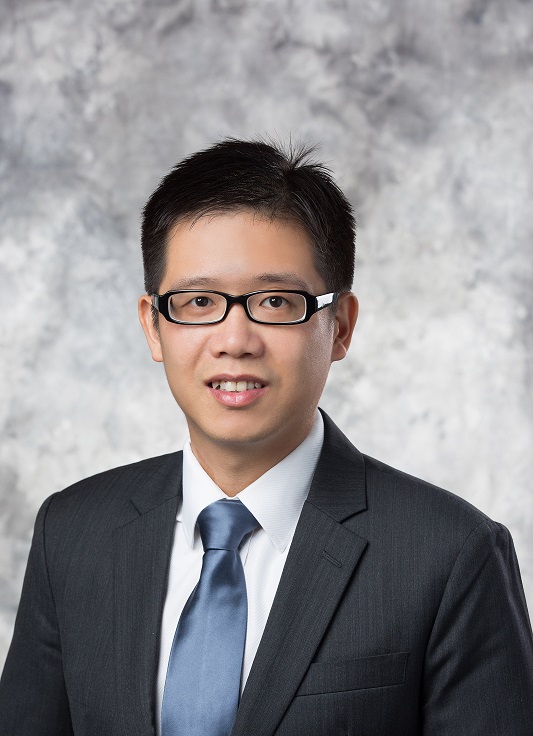}}]{Lok Ming Lui} received the Ph.D. degree in applied mathematics from the University of California at Los Angeles, Los Angeles, CA, USA, in 2008. He is an Associate Professor with the Department of Mathematics, The Chinese University of Hong Kong (CUHK), Hong Kong. Before joining CUHK, he was a Post-Doctoral Scholar for 2 years with the Department of Mathematics, Harvard University, Cambridge, MA, USA. His current research interests include computational conformal and quasi-conformal geometry, Teichmuller theory, surface registration, medical imaging, and shape analysis.
\end{IEEEbiography}


\vspace{-4mm}
\begin{thebibliography}{}

\bibitem{Avidan2007}
Shai Avidan and Ariel Shamir,
{\it Seam carving for content-aware image resizing}. 
ACM Transactions on graphics (TOG), 26(3), 2007.

\bibitem{Wolf2007}
Lior Wolf, Moshe Guttmann and Daniel Cohen-Or, 
{\it Non-homogeneous content-driven video-retargeting.} 2007 IEEE 11th International Conference on Computer Vision, 1--6, 2007.

\bibitem{Guo2009}
Yanwen Guo, Feng Liu, Jian Shi, Zhi-Hua Zhou, and Michael Gleicher, 
{\it Image retargeting using mesh parametrization.}
IEEE Transactions on Multimedia, 11(5), 856--867, 2009.

\bibitem{Zhang2009}
Guo-Xin Zhang, Ming-Ming Cheng, Shi-Min Hu, and Ralph R. Martin, 
{\it A Shape-Preserving Approach to Image Resizing.} 
Computer Graphics Forum, 28(7), 1897--1906, 2009.

\bibitem{NHO2010}
Yong Jin, Ligang Liu, and Qingbiao Wu, 
{\it Nonhomogeneous scaling optimization for realtime image resizing.} 
The Visual Computer, 26(6), 769--778, 2010.

\bibitem{Chen2010}
Renjie Chen, Daniel Freedman, Zachi Karni, Craig Gotsman, and Ligang Liu, 
{\it Content-aware image resizing by quadratic programming.} 
Computer Vision and Pattern Recognition Workshops, 1--8, 2010.

\bibitem{Panozzo2012}
Daniele Panozzo, Ofir Weber, and Olga Sorkine, 
{\it Robust Image Retargeting via Axis-Aligned Deformation.} 
Computer Graphics Forum, 31(2), 229--236, 2012.

\bibitem{Xu2017}
Jinlan Xu, Hongmei Kang, and Falai Chen,
{\it Content-aware image resizing using quasi-conformal mapping.} The Visual Computer, 1--12, 2017.

\bibitem{survey2010}
Daniel Vaquero, Matthew Turk, Kari Pulli, Marius Tico, and Natasha Gelfand, 
{\it A survey of image retargeting techniques.} 
Proc. SPIE, 7798, 1--15, 2010.

\bibitem{Lui2013}
Lok Ming Lui, Ka Chun Lam, Tsz Wai Wong, and Xianfeng Gu,
{\it Texture map and video compression using Beltrami representation}.
SIAM Journal on Imaging Sciences, 6(4), 1880--1902, 2013.

\bibitem{Lam14}
Ka Chun Lam, and Lok Ming Lui,
{\it Landmark and intensity based registration with large deformations via quasi-conformal maps}.
SIAM Journal on Imaging Sciences, 7(4), 2364--2392, 2014.

\bibitem{Wang2008}
Yu-Shuen Wang, Chiew-Lan Tai, Olga Sorkine, and Tong-Yee Lee, 
{\it Optimized scale-and-stretch for image resizing.} 
ACM Transactions on Graphics (TOG), 27(5), 2008.

\bibitem{Rubinstein2008}
Michael Rubinstein, Ariel Shamir, and Shai Avidan, 
{\it Improved seam carving for video retargeting.} 
ACM transactions on graphics (TOG), 27(3), 2008.

\bibitem{Rubinstein2009}
Michael Rubinstein, Ariel Shamir, and Shai Avidan, 
{\it Multi-operator media retargeting.} 
ACM Transactions on Graphics (TOG), 28(3), 2009.

\bibitem{Pritch09}
Yael Pritch, Eitam Kav-Venaki, and Shmuel Peleg, 
{\it Shift-map image editing.} 
2009 IEEE 12th International Conference on Computer Vision, 151--158, 2009.

\bibitem{Krähenbühl2009}
Philipp Krähenbühl, Manuel Lang, Alexander Hornung, and Markus Gross, 
{\it A system for retargeting of streaming video.} 
ACM Transactions on Graphics (TOG), 28(5), 2009.

\bibitem{Karni2009}
Zachi Karni, Daniel Freedman, and Craig Gotsman, 
{\it Energy-Based Image Deformation.} 
Computer Graphics Forum, 28(5), 1257--1268, 2009.

\bibitem{Rubinstein2010}
Michael Rubinstein, Diego Gutierrez, Olga Sorkine, and Ariel Shamir, 
{\it A comparative study of image retargeting.} 
ACM transactions on graphics (TOG), 29(6), 2010.

%\bibitem{Tillett05}
%J. Tillett, T.M. Rao, F. Sahin, R. Rao and S. Brockport,
%{\it Darwinian Particle Swarm Optimization}.
%Proceedings of the 2nd Indian international conference on artificial intelligence, pp. 1474--1487, 2005.

%\bibitem{Ballard81}
%D.H.Ballard,
%{\it Generalizing the Houghtransform to detectarbitraryshapes}.
%Pattern Recognition 13 (2): 111--122, 1981.
\end{thebibliography}
\end{document}